%% file: neurips_2026.tex
\documentclass{article}

 \usepackage[main, final]{neurips_2026}
\usepackage[utf8]{inputenc} % allow utf-8 input
\usepackage[T1]{fontenc}    % use 8-bit T1 fonts
\usepackage{hyperref}       % hyperlinks
\usepackage{url}            % simple URL typesetting
\usepackage{booktabs}       % professional-quality tables
\usepackage{amsfonts}       % blackboard math symbols
\usepackage{nicefrac}       % compact symbols for 1/2, etc.
\usepackage{microtype}      % microtypography
\usepackage{xcolor}         % colors
\usepackage{graphicx}
\usepackage{amssymb}
\usepackage{amsmath}
\usepackage{subcaption}
\usepackage{amsthm}

\usepackage{colortbl}
\usepackage{xcolor}
\definecolor{topcolor}{RGB}{198,239,206}   % green，top-1
\definecolor{secondcolor}{RGB}{189,215,238} % blue，top-2
\usepackage{pifont}
\newcommand{\cmark}{\ding{51}}
\newcommand{\xmark}{\ding{55}}
\usepackage{multicol}
\usepackage{multirow}
\usepackage{makecell}
\title{NormLift: From Lifted Features To Semantic Reliability In 3D Gaussian Splatting}

\author{%
  Yihan Zang \quad Da Li \quad Dominik Engel \quad Shinkyu Park \quad Ivan Viola \\
  CEMSE Division, King Abdullah University of Science and Technology \\
  \texttt{\{yihan.zang, da.li, dominik.engel, shinkyu.park, ivan.viola\}@kaust.edu.sa}
}
\begin{document}

\maketitle

\begin{abstract}
Training-free weighted aggregation is widely used to lift 2D semantic features onto 3D Gaussians for open-vocabulary scene understanding, yet its theoretical role remains insufficiently understood.
Existing analyses typically justify this operation from the rendering side, treating Gaussian features as linearly composable Euclidean variables for reconstructing 2D feature maps.
However, this view does not match downstream 3D usage, where each Gaussian is often queried independently in a cosine-based embedding space.
We revisit feature lifting from the 3D side and formulate per-Gaussian assignment as a cosine alignment problem on the CLIP unit sphere.
Under this objective, the $\ell_2$-normalized semantic back-projected feature emerges as the closed-form solution, providing a complementary interpretation of the standard lifting rule from the perspective of per-Gaussian semantic assignment.
This solution is still a convex combination of the observed directions, so every lifted feature is inherently a mixture. Such mixing is unavoidable for any per-Gaussian estimator, and the question is whether it can be measured.
The same formulation answers this: the norm of the back-projected feature decomposes into intra-view and inter-view consistency, so the feature magnitude quantifies how mixed the observations are and serves as a semantic reliability signal.
Calibrated by effective multi-view support, this reliability score identifies the Gaussians whose lifted direction has drifted and guides a mode-voting refinement that corrects them by copying a single observed CLIP direction instead of averaging.
Experiments on open-vocabulary 3D semantic segmentation show that NormLift is an efficient, training-free framework that achieves strong performance across evaluation protocols. Our code is available on \href{https://github.com/kaust-violab/NormLift_release}{GitHub}.
\end{abstract}

\section{Introduction}
\label{sec:intro}

3D Gaussian Splatting (3DGS)~\citep{kerbl20233dgs} has recently
emerged as an efficient and high-quality representation for
3D scene reconstruction~\citep{Wang_2026_CVPR}, with broad applications in generation,
editing, and downstream 3D tasks~\citep{dreamgaussian,gaussiandreamer,
gsdrag,wu2024recent}. Building on this representation, a
growing line of work extends 3DGS for open-vocabulary 3D
scene understanding by lifting 2D foundation-model
features, most prominently CLIP~\citep{radford2021clip}, onto
the 3D Gaussian primitives~\citep{semanticgaussians,goi,cags,alegret2026gala,identlgs}. The dominant strategy is a simple weighted aggregation: each Gaussian
receives the average of the 2D features it contributes to, weighted by
its alpha-blending contribution along each pixel ray. This operation,
introduced into the 3DGS literature by~\citet{joseph2025gradient} as
\emph{feature back-projection}, mirrors the classical back-projection
operation in computed tomography, which redistributes projection
measurements back into the image domain during
reconstruction~\citep{kak1988principles,natterer2001mathematics}.
Following this terminology, we refer to its CLIP-feature instantiation
as \emph{semantic back-projection}.

Despite its widespread adoption, the role of back-projection has been
analyzed almost from the rendering side, where the formula
arises as an approximate solution to a 2D feature reconstruction
objective, derived under simplifying assumptions or characterized by
bounded approximation guarantees. In this view, each Gaussian feature is
treated as a Euclidean component to be linearly composed through alpha
blending. However, this is not how Gaussians are typically used in
downstream 3D tasks: each is queried independently as a semantic
primitive on the CLIP feature space, where similarity is measured by
cosine on the unit sphere rather than by Euclidean reconstruction error.
This mismatch motivates a complementary, 3D-side perspective on the same
lifting operation, one that reflects the geometry of CLIP and the way
each Gaussian is consumed downstream.

Fixing the lifting rule by this objective does not remove a structural property of
per-Gaussian lifting. The solution is a convex combination of the observed CLIP
directions, so every lifted feature is inherently a mixture of what the Gaussian has
seen. This mixing is not a defect of back-projection in particular. A Gaussian is
observed by hundreds of rays across many views, and any representation that registers
a single feature onto it must aggregate them. The question is therefore not whether
mixing occurs, but whether its degree can be measured. Two kinds of
inconsistency dominate in practice: the same Gaussian may receive
conflicting features within a single view (\emph{intra-view
inconsistency}) or across different views (\emph{inter-view
inconsistency}). The same derivation yields an algebraic identity that
factors the norm of the back-projected feature into two terms in
$[0,1]$, capturing intra-view concentration and inter-view agreement,
respectively. This makes the norm itself a structural per-Gaussian indicator of how mixed the observations are, obtained from the lifting formulation
rather than from external visibility heuristics. Calibrated by an
effective view count to discount single-view dominance, the norm yields
a per-Gaussian \emph{reliability score} that, in our experiments,
aligns monotonically with downstream accuracy and captures information
largely complementary to opacity (Sec.~\ref{sec:exp3d},
Appendix~\ref{app:reliability-opacity}).

Experiments on ScanNet show that NormLift improves over 
prior training-free and training-based baselines on open-vocabulary 3D 
semantic segmentation. The reliability score is largely independent of 
opacity except in a small zero-evidence regime, capturing a per-Gaussian 
signal beyond what visibility-based heuristics provide. Its 
reliability-guided KNN refinement is lightweight, running 
$\mathbf{6.7\times}$ faster than SFS's post-lifting pipeline with 
matched peak memory. The lifted features further transfer to a 2D 
rendering-based protocol on LERF-OVS, where NormLift attains the best 
mean localization accuracy. In summary, our contributions are:

\begin{itemize}
  \item We formulate per-Gaussian feature lifting as a cosine alignment
  problem on the CLIP unit sphere and show that the $\ell_2$-normalized
  back-projected feature is its closed-form optimum. This complements
  existing rendering-side analyses with a 3D-side interpretation that
  reflects the cosine geometry.
  
\item We show that the inherent mixing of per-Gaussian lifting is measurable: the
  back-projected norm factors algebraically into intra-view and inter-view
  consistency, and calibrated by effective multi-view support, yields a per-Gaussian
  reliability score R(j) that flags Gaussians whose lifted direction has drifted away
  from any single observed semantics. R(j) is largely independent of opacity outside
  a small zero-evidence regime.
  
  \item Guided by $R(j)$, we design a mode-voting refinement that corrects unreliable Gaussians by
  copying a single observed CLIP direction from spatial neighbors rather
  than linearly averaging them, in accordance with the non-linearity of
  the CLIP semantic manifold.
  
  \item NormLift is training-free, improves over prior 
  training-free and training-based baselines on ScanNet, and runs $\mathbf{6.7\times}$ faster than SFS in 
  the post-lifting stage at matched peak memory.
\end{itemize}

\section{Related Work}
\label{sec:related}

\subsection{Open-vocabulary 3D understanding with Gaussian splatting}

Existing open-vocabulary 3DGS methods can be broadly grouped into
training-based, grouping-based, and training-free approaches.
Training-based methods learn semantic Gaussian features through
rendering-based supervision, including LangSplat~\citep{qin2024langsplat},
LEGaussians~\citep{legaussians}, Feature 3DGS~\citep{feature3dgs},
N2F2~\citep{bhalgat2024n2f2}, and GAGS~\citep{peng2026gags}. These methods
distill CLIP or other foundation-model features into Gaussian
representations, often using scene-specific compression, quantization,
or granularity-aware supervision to improve 2D open-vocabulary
rendering. Grouping-based methods instead introduce object-, instance-,
or hierarchy-level structures to improve 3D semantic consistency,
including OpenGaussian~\citep{opengaussian},
GaussianGraph~\citep{wang2025gaussiangraph},
SuperGSeg~\citep{liang2026supergseg},
VoteSplat~\citep{jiang2025votesplat}, LaGa~\citep{laga}, and
THGS~\citep{thgs}. These methods are effective at producing structured
semantic representations, but typically depend on per-scene
optimization, learned embeddings, or scene-level grouping stages.

A recent line of training-free methods removes gradient-based semantic
optimization and directly transfers 2D features to Gaussians using
rendering or visibility weights. Although these methods differ in their
implementation details, they largely revolve around the same core
operation: a rendering-weighted aggregation of 2D features onto 3D
Gaussians, or a variant of this operation.
Gradient-Weighted Back-Projection~\citep{joseph2025gradient}, from which we adopt the back-projection terminology, performs a fast training-free lifting via gradient-based contribution weights; LBG~\citep{chacko2025lifting} performs visibility-weighted lifting
without per-scene training; Occam's LGS~\citep{occams} derives a
probabilistic weighted aggregation rule; SFS~\citep{sfs} formulates
feature lifting as a sparse linear inverse problem with a closed-form
solution under a global $L_2$ reconstruction objective;
LUDVIG~\citep{marrie2025ludvig} interprets lifting as inverse rendering and
further refines features with DINOv2-based graph diffusion;
Dr.Splat~\citep{jun2025dr} registers top-$k$ Gaussians per ray with
product quantization for compact storage; and
VALA~\citep{wang2026visibility} introduces visibility-aware gating with
a cosine geometric median for multi-view aggregation. A common feature
of these analyses is that Gaussian features are treated as Euclidean
variables that are linearly composed through alpha blending to
reproduce 2D feature maps, which leads to approximate, bounded, or
post-refined formulations on the rendering side.

Our work belongs to this training-free family, but studies the same
lifting operation from a complementary perspective. Instead of asking
which Gaussian features best reconstruct the 2D feature maps, we ask,
for each Gaussian independently, which unit direction on the CLIP
sphere is most supported by the 2D observations it contributes to.
Under this 3D-side formulation, the $\ell_2$-normalized back-projected
feature appears directly as the closed-form optimum, and unit
normalization is handled by the objective itself rather than as a
post-processing step. This per-Gaussian view is complementary to rendering-side analyses and matches the cosine-based geometry under which lifted features are actually queried downstream.

\subsection{Per-Gaussian reliability and filtering}

Several recent methods use reliability- or visibility-related cues to
stabilize semantic Gaussians. Occam's LGS~\citep{occams} filters
Gaussians with negligible rendering contributions;
VALA~\citep{wang2026visibility} uses marginal ray contributions to gate
visible Gaussians and aggregates multi-view features with a cosine
geometric median; and ReLaGS~\citep{Xie_2026_CVPR} removes
low-contribution Gaussians through maximum-weight pruning. Other
methods improve robustness through feature assignment or grouping
mechanisms: Dr.Splat~\citep{jun2025dr} registers CLIP features to
dominant Gaussians along each ray with product quantization,
OpenGaussian~\citep{opengaussian} uses codebook-based feature
discretization and instance-level 3D--2D association, and
LaGa~\citep{laga} builds object-level view-aggregated descriptors after
scene decomposition.

These designs are effective but obtain their reliability cues
\emph{externally} to the lifting operation, from visibility statistics,
rendering contributions, feature assignment, or object-level grouping.
The closest in spirit is VALA, which performs robust multi-view
aggregation via a cosine geometric median; this can be viewed as
defining a separate aggregation rule on top of the lifted features. In
contrast, our reliability signal is obtained \emph{internally} from the
same lifting formulation: the norm of the back-projected feature
algebraically factors into intra-view and inter-view consistency, and
is then calibrated by effective multi-view support. The lifted
direction $u_j^\star = f_j/\|f_j\|$ and the reliability score $R(j)$
are therefore two outputs of a single per-Gaussian objective, rather
than the result of two separate stages. We further show empirically
that $R(j)$ overlaps with opacity-based signals only at the
zero-evidence boundary and is largely independent of opacity for the
remaining $84\%$ of Gaussians
(Sec.~\ref{sec:exp3d}, Appendix~\ref{app:reliability-opacity}),
indicating that it provides information complementary to existing
visibility- and rendering-based cues.

\section{Method}

\begin{figure}[t]
  \centering
  \includegraphics[width=\linewidth]{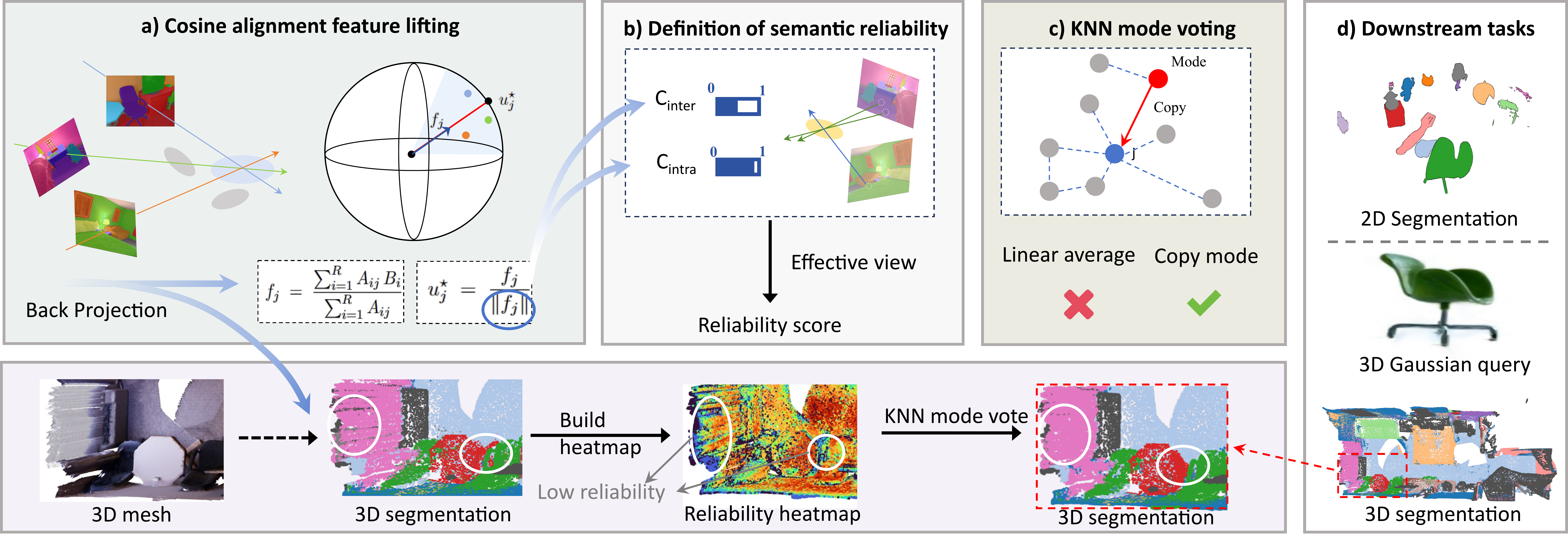}
\caption{Method overview of NormLift. \textbf{Top}: a) Per-Gaussian
features are lifted via cosine alignment on the CLIP sphere,
giving the closed-form $u_j^\star = f_j/\|f_j\|$. b) The norm
$\|f_j\|$ factors into intra- and inter-view consistency,
calibrated to the reliability score $R(j)$ by effective view support. c) KNN mode voting
copies the most-supported neighbor's feature, avoiding the blur
of linear averaging. d) NormLift supports 2D and 3D downstream tasks. \textbf{Bottom}: a visualization of the 3D segmentation pipeline. The framework is training-free and operates entirely at the per-Gaussian level.}
  \label{fig:me}
  \vspace{-3mm}
\end{figure}

As illustrated in Figure~\ref{fig:me}, this section recasts the
widely-used back-projection heuristic as the closed-form optimum of a
per-Gaussian cosine alignment problem, and shows that the norm of the
back-projected feature, calibrated by effective view support, yields a
reliability score for KNN mode voting refinement.
Section~\ref{sec:prelim} revisits the standard feature lifting setup
and distills three geometric properties of CLIP features.
Section~\ref{sec:cosine} formulates per-Gaussian lifting as a
constrained optimization on the CLIP unit sphere, whose closed-form
solution coincides with the $\ell_2$-normalized back-projected feature.
Section~\ref{sec:reliability} shows that the same norm admits an
algebraic decomposition into intra- and inter-view consistency,
yielding the optimal direction (its unit vector) and a reliability
measure (its norm) simultaneously; we calibrate this norm by effective
multi-view support to obtain $R(j)$.
Section~\ref{sec:refinement} then leverages $R(j)$ in a mode-voting
procedure to refine unreliable Gaussians.

\subsection{Preliminary}
\label{sec:prelim}

\subsubsection{3D Gaussian splatting}
\label{sec:splat-rendering}

A 3D Gaussian Splatting 
(3DGS)~\citep{kerbl20233dgs} scene is represented by a set of 3D Gaussian primitives, which are 
parameterized by their position, covariance, opacity, and color. Images are rendered through a fixed 
depth-sorted alpha-blending pipeline.

Although 3D Gaussian
Splatting is implemented via screen-space rasterization, its alpha-compositing process can be interpreted in a ray-centric manner for analysis: each pixel corresponds
to a viewing ray, and the Gaussians
projected onto that pixel can be ordered by depth and composited in a
front-to-back manner. This view is only used as an analytical
description of the rasterized compositing process, and does not imply
that standard 3DGS performs ray casting.

Consider all pixel observations in the training views, indexed by
$i \in \{1,\dots,R\}$. For each pixel/ray $i$, let the contributing
Gaussians be ordered according to their depth along the corresponding
camera direction. The contribution weight of Gaussian $j$ to observation
$i$ is given by the standard alpha-compositing rule:

\vspace{-2em}
\begin{equation}
  A_{ij} \;=\; \alpha_{ij} \prod_{k < j} \bigl(1 - \alpha_{ik}\bigr),
  \label{eq:alpha-blend}
\end{equation}
where $\alpha_{ij} \in [0, 1]$ denotes the opacity contribution of 
Gaussian $j$ to ray $i$, and $k<j$ indicates Gaussians
that are closer to the camera in the same compositing order. The rendered
color can then be written as

\vspace{-2.5em}
\begin{equation}
  C_i \;=\; \sum_{j=1}^{P} A_{ij}\, c_j 
  \;+\; \Bigl(1 - \sum_{j=1}^{P} A_{ij}\Bigr)\, C_b,
  \label{eq:render}
\end{equation}
where $c_j$ is the color attribute of Gaussian $j$ in standard 3DGS and
$C_b$ is the background color. In this work, we use the same compositing
weights $A_{ij}$ as the bridge between 2D semantic observations and 3D
Gaussian primitives. In particular, while $c_j$ denotes RGB color in
standard rendering, the same weighted compositing can be applied to
a semantic feature attribute when analyzing feature lifting.

\subsubsection{Back-projection}
\label{sec:back-projection}

Given a trained 3DGS scene, the goal of feature lifting is to assign 
each Gaussian $j$ a semantic feature $u_j \in \mathbb{R}^F$ that 
supports downstream tasks such as open-vocabulary segmentation 
and querying. The supervision comes from 2D semantic 
observations $\{B_i\}_{i=1}^{R}$, one per rendering ray, obtained by 
applying a foundation model such as CLIP~\citep{radford2021clip} to 
the rendered views. In this paper, we concentrate on the CLIP features, which are normalized, in line with the unit spherical 
geometry.

The prototype of the widely used training-free strategy is to aggregate the 2D observations 
onto each Gaussian using the alpha-blending weights from 
Eq.~\eqref{eq:alpha-blend}. Concretely, for each Gaussian $j$, one 
forms the rendering-weighted average
\begin{equation}
  f_j \;=\; \frac{\sum_{i=1}^{R} A_{ij}\, B_i}{\sum_{i=1}^{R} A_{ij}},
  \label{eq:backproj}
\end{equation}
which is typically further $\ell_2$-normalized for downstream 
use, in line with the spherical geometry of CLIP.

\subsubsection{Geometric properties of CLIP features}
\label{sec:clip-properties}
Before formulating feature lifting from the 3D perspective, we first clarify three geometric properties of CLIP-like embeddings that are essential for analyzing the behavior of lifted Gaussian features. These follow from how CLIP is trained
and consumed; they are objective characteristics of the feature
space, not assumptions we impose.

\paragraph{Property~1 Sphere constraint:}
CLIP features lie on the unit hypersphere $S^{F-1}$, since both image
and text encoders are $\ell_2$-normalized before the contrastive
objective~\citep{radford2021clip}. Any lifted feature $u_j$ compared
against CLIP queries must satisfy $\|u_j\| = 1$.

\paragraph{Property~2 Directional semantics:}
CLIP is trained with a cosine objective. Semantic similarity is measured by cosine, not Euclidean distance.

\begin{figure}[t]
  \centering

  \begin{subfigure}[t]{0.48\linewidth}
    \centering
    \raisebox{0.28cm}{%
      \includegraphics[width=\linewidth]{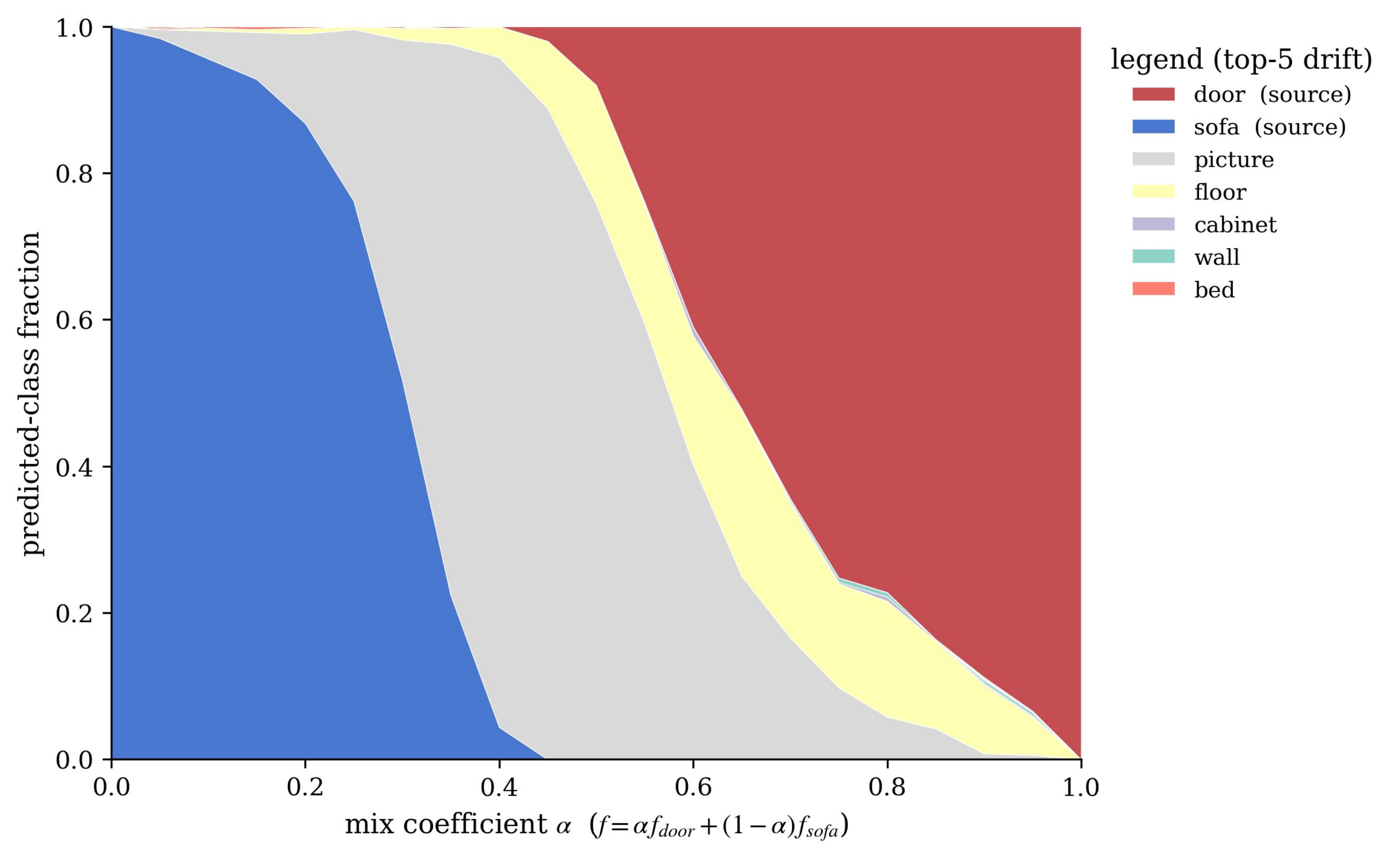}
    }
    \label{fig:sofa-door-drift}
  \end{subfigure}
  \hspace{0.02\linewidth}
  \begin{subfigure}[t]{0.38\linewidth}
    \centering
    \includegraphics[width=\linewidth]{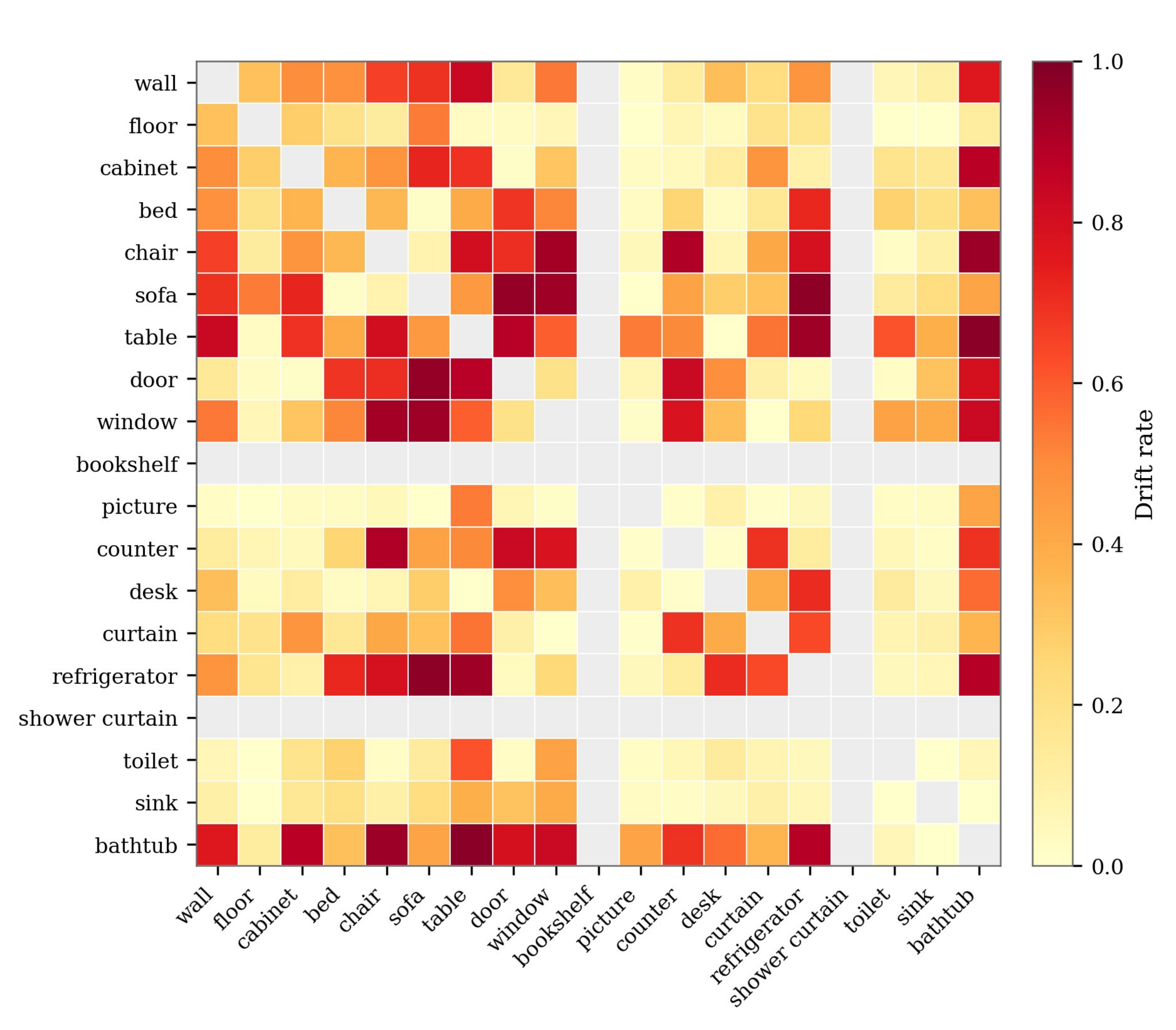}
    \label{fig:pairwise-drift-rate}
  \end{subfigure}
  \vspace{-1.2em}

\caption{\small
Semantic drift from linear interpolation of CLIP features.
\textbf{Left:} mixing the source features of \emph{door} and \emph{sofa} 
as $f=\alpha f_{\mathrm{door}}+(1-\alpha)f_{\mathrm{sofa}}$. The 
predicted class drifts to unrelated categories (e.g.\ \emph{picture}, 
\emph{floor}) for $\alpha\in[0.3,0.7]$, with neither source class 
dominating in this regime.
\textbf{Right:} pairwise drift rate at $\alpha{=}0.5$ across the 19 
ScanNet classes used in our protocol; the overall drift rate is 
$33.6\%$.
}
  \label{fig:clip-linear-drift}
  \vspace{-1.2em}
\end{figure}

\paragraph{Property~3 Non-linearity of the semantic manifold:}
The CLIP semantic manifold is not globally closed under linear combination. Small-angle mixtures may remain locally coherent, but mixtures of semantically distant features may induce semantic drift. Figure~\ref{fig:clip-linear-drift} shows that a linear combination of CLIP features from two distinct objects may lead to semantic drift.

\subsection{Per-Gaussian cosine alignment problem}
\label{sec:cosine}

\subsubsection{Problem formulation}
\label{sec:cosine-formulation}

Building on the properties of CLIP features, we formulate per-Gaussian feature
lifting as an optimization problem on the CLIP unit sphere. 
Let $u_j \in S^{F-1}$ denote the semantic feature to be assigned to
Gaussian $j$. We define the per-Gaussian cosine alignment problem as
\vspace{-0.5em}
\begin{equation}
  u_j^* \;=\; \arg\max_{u \in S^{F-1}}
  \sum_{i=1}^{R} A_{ij}\, \langle u, B_i \rangle,
  \qquad j = 1, \dots, P.
  \label{eq:cosine-alignment}
\end{equation}
\vspace{-1.5em}

Eq.~\eqref{eq:cosine-alignment} defines a contribution-weighted
semantic consensus for each Gaussian. For a fixed Gaussian $j$, each
observation contributes a scalar agreement
$\langle u,B_i\rangle$ between the candidate Gaussian feature $u$ and
the 2D CLIP direction $B_i$, weighted by the alpha-compositing
contribution $A_{ij}$. Thus, the objective averages semantic
agreements rather than assuming that CLIP features themselves are
physically composable through rendering. The linear combination that
appears below is only a consequence of the linearity of the inner
product, not an assumption that mixed CLIP features form a valid
semantic feature. This gives a 3D-side view of lifting: each Gaussian
is treated as an independent semantic primitive whose unit feature is
chosen to best agree with its contribution-weighted observations.

By linearity of the inner product, the objective can be rewritten as
$\max_{\|u\|=1}\langle u, f_j\rangle$, where $f_j$ is the
back-projected feature in Eq.~\eqref{eq:backproj}. For $f_j\neq 0$,
Cauchy--Schwarz gives

\vspace{-1em}
\begin{equation}
  u_j^\star = \frac{f_j}{\|f_j\|},
  \qquad
  \text{with optimal value } \|f_j\| .
  \label{eq:cosine-optimum}
\end{equation}
\vspace{-1em}

We note that $f_j$ is weight-normalized (divided by $\sum_i A_{ij}$)
but not unit-normalized; as a convex combination of unit vectors, it
satisfies $\|f_j\|\in[0,1]$, a property used in
Sec.~\ref{sec:reliability}. Appendix~\ref{app:closed-form-derivation} shows the full three-step derivation, together
with the treatment of the boundary case $f_j=0$.

The derivation is simple; the key point is the objective it solves.
Under the per-Gaussian cosine-alignment view, the standard
back-projection rule followed by $\ell_2$ normalization is not merely a
post-processing heuristic, but the closed-form solution to a semantic
agreement objective on the CLIP sphere.

\subsubsection{What the back-projected norm measures}
\label{sec:cosine-reframing}

\paragraph{Where the residual error lives:}
Since $u_j^\star$ is the unique maximizer of
Eq.~\eqref{eq:cosine-alignment} for $f_j \neq 0$
(Appendix~\ref{app:closed-form-derivation}), the formulation itself
introduces no further error. Any mismatch between $u_j^\star$ and the
true semantic identity of Gaussian $j$ must therefore come from the
observations themselves: the 2D features $\{B_i\}$ may be noisy,
inconsistent across views, or contaminated by pixels that mix multiple
semantic regions. Equivalently, $f_j$ is a convex combination of the observed unit directions
$B_i$. When these directions disagree, $u_j^\star$ is their compromise and may
point toward a semantics that none of the observations individually carries.
Such mixing is inherent to any per-Gaussian estimator, since a single vector
must summarize many rays; what the formulation offers is a way to quantify it,
which we develop next.

\paragraph{The norm as a data-side consistency signal:}
The norm $\|f_j\|$ provides a natural handle on this observational
uncertainty. Because $u_j^\star = f_j/\|f_j\|$ already absorbs the
geometric content of the optimum, the magnitude $\|f_j\|$ is free to
encode something else: how strongly the contribution-weighted
observations agree with their own resultant direction. The next
section makes this precise by decomposing $\|f_j\|$ into intra-view
and inter-view consistency factors, which we then calibrate into a
per-Gaussian reliability score.

\subsection{Semantic reliability score}
\label{sec:reliability}

\subsubsection{Norm decomposition: an algebraic identity}
\label{sec:norm-decomp}

We now make the decomposition explicit. The norm $\|f_j\|$ admits an
algebraic factorization into two terms in $[0,1]$, capturing
within-view and across-view consistency respectively. The factorization
follows directly from how $f_j$ aggregates per-view averages, and
should be read as an algebraic identity rather than a theorem.

Let $\mathcal{R}_v\subseteq\{1,\dots,R\}$ denote the set of pixel/ray
observations from view $v$. For Gaussian $j$, define the per-view
weight and per-view aggregation as

\vspace{-1.5em}
\begin{equation}
  W_j^{(v)} := \sum_{i\in\mathcal{R}_v} A_{ij},
  \qquad
  f_j^{(v)} :=
  \frac{\sum_{i\in\mathcal{R}_v} A_{ij} B_i}{W_j^{(v)}} .
  \label{eq:perview}
\end{equation}

The global feature can then be written as
$f_j = \bigl(\sum_v W_j^{(v)} f_j^{(v)}\bigr)/\bigl(\sum_v W_j^{(v)}\bigr)$.
Multiplying the numerator and denominator by
$\sum_v W_j^{(v)}\|f_j^{(v)}\|$ yields

\begin{equation}
  \|f_j\|
  =
  \frac{\left\|\sum_v W_j^{(v)} f_j^{(v)}\right\|}
       {\sum_v W_j^{(v)}}
  =
  \underbrace{
  \frac{\sum_v W_j^{(v)}\|f_j^{(v)}\|}
       {\sum_v W_j^{(v)}}
  }_{C_{\mathrm{intra}}(j)}
  \cdot
  \underbrace{
  \frac{\left\|\sum_v W_j^{(v)} f_j^{(v)}\right\|}
       {\sum_v W_j^{(v)}\|f_j^{(v)}\|}
  }_{C_{\mathrm{inter}}(j)} .
  \label{eq:decomp}
\end{equation}

The two factors separate complementary sources of consistency.
$C_{\mathrm{intra}}(j)$ captures within-view concentration, decreasing
when a view assigns mixed semantic directions to Gaussian $j$.
$C_{\mathrm{inter}}(j)$ captures cross-view agreement through a weighted
mean resultant length over the per-view directions
$\{f_j^{(v)}/\|f_j^{(v)}\|\}_v$~\citep{mardia2009directional}.
Therefore, $\|f_j\|$ is high only when the supporting observations agree
both within and across views, making it a structural indicator of
semantic consistency. Since a single dominant view can still yield a
high norm, we further calibrate it with effective multi-view support.

\subsubsection{Calibrating with effective views}
\label{sec:reliability-calibration}

The norm $\|f_j\|$ alone does not distinguish many agreeing views from
a single dominant one. To make this distinction explicit, we calibrate
$\|f_j\|$ by the classical effective sample
size~\citep{kish1965survey} applied to the per-view rendering weights,
$N_{\mathrm{eff}}(j) := \bigl(\sum_v W_j^{(v)}\bigr)^2 / \sum_v
\bigl(W_j^{(v)}\bigr)^2$, which equals $1$ when one view dominates and
grows toward the visible-view count when the weights are uniform.
This reflects how many views \emph{effectively} contribute rather than
how many merely see the Gaussian. We define the per-Gaussian
\emph{semantic reliability score} as
\begin{equation}
  R(j) \;:=\; \|f_j\| \cdot
  \frac{N_{\mathrm{eff}}(j)}{N_{\mathrm{eff}}(j) + \beta} .
  \label{eq:reliability}
\end{equation}
The shrinkage factor is a standard Bayesian mean-shrinkage with prior
strength $\beta$; we use $\beta=1$ (one ``virtual view'') and analyze
its sensitivity in Appendix~\ref{app:refinement-hparams}. $R(j)$ is
high only when the lifted feature is both directionally consistent and
backed by non-trivial multi-view evidence.

Appendix~\ref{app:reliability-basis} collects the geometric and
optimization arguments for reading $\|f_j\|$ as a reliability signal:
it is the mean resultant length of the observed directions and, at the
same time, the optimal value attained by Eq.~\eqref{eq:cosine-alignment}. Appendix~\ref{app:trend} shows $R(j)$ aligns more monotonically with downstream accuracy
than $\|f_j\|$ alone. Appendix~\ref{app:reliability-opacity} shows it also overlaps
with opacity mainly in the zero-evidence regime ($R(j)=0$, $\sim$$16\%$
of Gaussians on ScanNet) and is largely uncorrelated with opacity for
the remaining Gaussians, providing a signal complementary to opacity-
and visibility-based heuristics.

% \begin{figure}[t]
%     \centering
%     \includegraphics[width=1.0\linewidth]{figures/NEW_NEW_QUERY.pdf}
%     \vspace{-5mm}
%     \caption{Qualitative comparison with SFS~\citep{sfs}.Predicted 3D Gaussian primitives retrieved by
% text queries on LERF scenes (figurines on the top, teatime at the bottom). The queried
% regions are annotated with colored bounding boxes on the
% original images.}
%     \label{fig:placeholder}
%     \vspace{-3mm}
% \end{figure}

\subsection{Reliability-guided KNN mode-voting refinement}
\label{sec:refinement}

The reliability score $R(j)$ flags which Gaussians carry trustworthy
lifted semantics, leaving open how to correct the rest. The
non-linearity of the CLIP semantic manifold
(Section~\ref{sec:clip-properties}, Property~3) suggests a constraint:
linearly averaging neighboring CLIP features can drift off-manifold,
so refinement should instead \emph{copy} a single coherent direction
from nearby candidates rather than blend them.

\paragraph{Neighbor mode selection:}
For each Gaussian $i$, we form a candidate set
$\mathcal{N}_K^+(i):=\mathcal{N}_K(i)\cup\{i\}$ from its $K$ nearest
spatial neighbors and itself (KNN in 3D Euclidean space). Rather than
averaging their features, we select one candidate by a
reliability-weighted support score. For each
$j\in\mathcal{N}_K^+(i)$,

\vspace{-0.5em}
\begin{equation}
  S_{ij}
  =
  R(j)
  \sum_{k\in\mathcal{N}_K^+(i)}
  R(k)\, d_{ik}\, g_{jk},
  \label{eq:support}
\end{equation}
\vspace{-1em}

where $d_{ik}=\exp(-\|\mu_i-\mu_k\|^2/(2\sigma_d^2))$ is a spatial
decay and $g_{jk}=\sigma((\langle u_k^\star,u_j^\star\rangle-\tau)/\gamma)$
is a soft semantic-agreement gate. The score has a simple voting
structure: each neighbor $k$ casts a vote for candidate $j$, weighted
by its own reliability $R(k)$, its spatial closeness $d_{ik}$ to the
target $i$, and its semantic agreement $g_{jk}$ with $j$; the outer
factor $R(j)$ then scales by the candidate's own reliability. A
candidate is thus supported when reliable neighbors near $i$ agree
with it.

\paragraph{Conservative replacement:}
Let $j^\star=\arg\max_{j\in\mathcal{N}_K(i)} S_{ij}$ be the best
neighbor excluding $i$. We replace $u_i^\star$ only if a margin
condition holds:

\begin{equation}
  u_i^\star \leftarrow u_{j^\star}^\star
  \quad \text{iff} \quad
  S_{ij^\star} > S_{ii}+\Delta,
  \label{eq:margin}
\end{equation}

and keep $u_i^\star$ otherwise. The margin $\Delta$ guards against
overwriting features when neighborhood evidence is only marginally
stronger. The hyperparameters $(\sigma_d,\tau,\gamma,\Delta)$ are
fixed across all experiments; defaults and a sensitivity analysis are
provided in Appendix~\ref{app:refinement-hparams}. Because every
refined feature is an existing $u_k^\star$, the procedure stays on
$S^{F-1}$ by construction and never forms a CLIP direction by linear
combination.

\begin{table}[t]

  \caption{Open-vocabulary 3D semantic segmentation on ScanNet under
  the OpenGaussian~\citep{opengaussian} protocol. T-F denotes
  training-free methods. Best and second-best results are highlighted.}
  \label{tab:scannet-3d}
  \centering
  \footnotesize
  \setlength{\tabcolsep}{2.2pt}
  \renewcommand{\arraystretch}{0.95}
  \begin{tabular}{lc cc cc cc}
    \toprule
    & & \multicolumn{2}{c}{19 cls.}
      & \multicolumn{2}{c}{15 cls.}
      & \multicolumn{2}{c}{10 cls.} \\
    \cmidrule(lr){3-4} \cmidrule(lr){5-6} \cmidrule(lr){7-8}
    Method & T-F
      & mIoU & mAcc
      & mIoU & mAcc
      & mIoU & mAcc \\
    \midrule
    LangSplat~\citep{qin2024langsplat}
      & \xmark
      & 3.78 & 9.11
      & 5.35 & 13.20
      & 8.40 & 22.06 \\
    OpenGaussian~\citep{opengaussian}
      & \xmark
      & 24.73 & 41.54
      & 30.13 & 48.25
      & 38.29 & 55.19 \\
    LaGa~\citep{laga}
      & \xmark
      & 32.50 & 49.10
      & 35.50 & 53.50
      & 42.60 & 63.20 \\
    \midrule
    THGS~\citep{thgs}
      & \cmark
      & \cellcolor{secondcolor}34.39 & 50.74
      & \cellcolor{topcolor}39.61 & \cellcolor{secondcolor}57.07
      & 46.38 & 64.74 \\
    VALA~\citep{wang2026visibility}
      & \cmark
      & 32.11 & 50.05
      & 35.10 & 54.77
      & 46.21 & 65.61 \\
    Occam's LGS~\citep{occams}
      & \cmark
      & 31.93 & 48.93
      & 34.25 & 53.71
      & 45.16 & 64.39 \\
    SFS~\citep{sfs}
      & \cmark
      & 33.33 & 51.35
      & 36.43 & 55.38
      & 44.74 & 63.53 \\
    LUDVIG~\citep{marrie2025ludvig}
      & \cmark
      & 33.90 & \cellcolor{secondcolor}51.40
      & 37.40 & 57.20
      & \cellcolor{secondcolor}46.40 & \cellcolor{secondcolor}66.20 \\
    \midrule
    \textbf{NormLift}
      & \cmark
      & \cellcolor{topcolor}35.82
      & \cellcolor{topcolor}54.15
      & \cellcolor{secondcolor}39.38
      & \cellcolor{topcolor}59.27
      & \cellcolor{topcolor}49.61
      & \cellcolor{topcolor}69.41 \\
    \bottomrule
  \end{tabular}

\end{table}

\begin{figure}[t]
  \centering
  \hspace*{-6mm}\includegraphics[width=\linewidth]{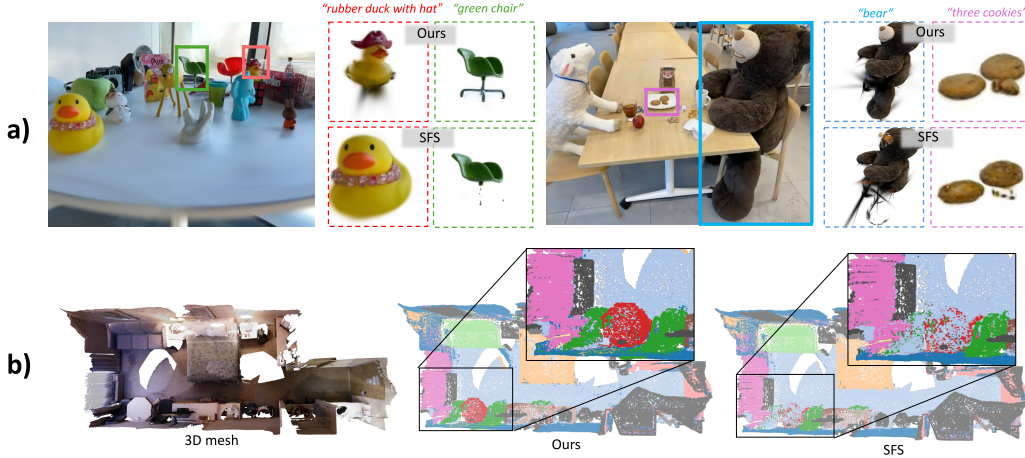}
  \vspace{-4mm}
\caption{Qualitative comparison with SFS~\citep{sfs} under the same
training-free protocol. NormLift improves text-query retrieval on
LERF scenes and produces sharper, more complete 3D segmentation on
ScanNet, especially in the zoomed-in regions.}
  \label{fig:qual-comparison}

\end{figure}

\section{Experiment}

In this section, we evaluate NormLift. Section~\ref{sec:exp3d} reports our main results on
open-vocabulary 3D semantic segmentation on ScanNet with each Gaussian queried as an
independent 3D semantic primitive. 
Section~\ref{sec:ablation} ablates each component of the
reliability score and refinement procedure. We further evaluate 2D rendering on 
LERF-OVS in Appendix~\ref{sec:exp2d}: although the 2D protocol 
composites Gaussians rather than querying them independently, our 
formulation remains competitive, indicating reasonable 2D 
generalization. All the experiments were completed on one NVIDIA Tesla V100 GPU and one NVIDIA RTX 4090 GPU.

\subsection{Open-vocabulary 3D experiments}
\label{sec:exp3d}

We evaluate NormLift on ScanNet~\citep{dai2017scannet} for
open-vocabulary 3D semantic segmentation, following the OpenGaussian protocol
~\citep{opengaussian} (Appendix~\ref{app:scannet-protocol}).

Table~\ref{tab:scannet-3d} compares NormLift with both
training-based language-field methods
and recent
training-free lifting methods.
NormLift achieves the best mIoU and mAcc, outperforming both
families. 
Visual comparisons in Figure~\ref{fig:qual-comparison} show
that NormLift produces more complete object boundaries and more details than SFS~\citep{sfs}, particularly on
thin structures and Gaussians with limited multi-view coverage,
consistent with the behavior expected from the reliability score
and refinement procedure.

\begin{table}[t]
\centering
\begin{minipage}[t]{0.50\linewidth}
\centering
\caption{Ablation of NormLift components on ScanNet under the
OpenGaussian protocol with 19/15/10 label granularities. Each variant
changes one component while keeping the others fixed. Reliability-weighted
mode voting gives the strongest result.}
\label{tab:ablation}
\footnotesize
\setlength{\tabcolsep}{3pt}
\renewcommand{\arraystretch}{1.08}
\resizebox{\linewidth}{!}{%
\begin{tabular}{l cc cc cc}
\toprule
& \multicolumn{2}{c}{19 cls.}
& \multicolumn{2}{c}{15 cls.}
& \multicolumn{2}{c}{10 cls.} \\
\cmidrule(lr){2-3} \cmidrule(lr){4-5} \cmidrule(lr){6-7}
Variant
  & mIoU & mAcc
  & mIoU & mAcc
  & mIoU & mAcc \\
\midrule
w/o refinement
  & 32.94 & 50.45 & 36.30 & 55.57 & 45.34 & 64.18 \\
plain KNN voting
  & 34.06 & 51.83 & 38.07 & 57.79 & 46.94 & 66.12 \\
linear averaging
  & 34.78 & 52.36 & 39.27 & 58.73 & 48.50 & 67.21 \\
norm-only reliability
  & 32.87 & 50.41 & 36.26 & 55.24 & 45.08 & 64.10 \\
support-only reliability
  & 33.67 & 51.64 & 37.95 & 57.20 & 47.02 & 65.79 \\
\midrule
\textbf{Full NormLift}
  & \textbf{35.82} & \textbf{54.15}
  & \textbf{39.38} & \textbf{59.27}
  & \textbf{49.61} & \textbf{69.41} \\
\bottomrule
\end{tabular}}
\end{minipage}%
\hfill
\begin{minipage}[t]{0.48\linewidth}
\centering
\caption{Wall-clock time and peak GPU memory on LERF-OVS using an NVIDIA Tesla V100 GPU. NormLift and SFS share the same lifting stage, while NormLift is
faster in post-lifting and reduces end-to-end runtime with similar
memory use.}
\label{tab:timing}
\footnotesize
\setlength{\tabcolsep}{3pt}
\renewcommand{\arraystretch}{1.25}
\resizebox{\linewidth}{!}{%
\begin{tabular}{l c | c c | c c | c}
\toprule
& \multirow{2}{*}{Lift (s)}
& \multicolumn{2}{c|}{Post-Lift (s)}
& \multicolumn{2}{c|}{Total (s)}
& \multirow{2}{*}{Mem (GB)} \\
\cmidrule(lr){3-4}\cmidrule(lr){5-6}
Scene & & SFS & \textbf{Ours} & SFS & \textbf{Ours} & \\
\midrule
Figurines & 449.2 & 379.5 & \textbf{34.8}  & 828.8 & \textbf{484.0} &  7.48 \\
Ramen     & 131.0 & 117.9 & \textbf{10.8}  & 248.9 & \textbf{141.8} &  4.71 \\
Teatime   & 189.5 & 532.3 & \textbf{101.2} & 721.8 & \textbf{290.7} & 15.17 \\
Waldo     & 200.3 & 406.1 & \textbf{67.0}  & 606.3 & \textbf{267.3} & 12.09 \\
\midrule
Mean & 242.5 & 359.0 & \textbf{53.5} & 601.5 & \textbf{295.9} & 9.86 \\
\textbf{Speedup} & --- & \multicolumn{2}{c|}{$\mathbf{6.7\times}$} & \multicolumn{2}{c|}{$\mathbf{2.0\times}$} & --- \\
\bottomrule
\end{tabular}}
\end{minipage}
\vspace{-4mm}
\end{table}

\subsection{Ablation study}\label{sec:ablation}

\paragraph{Component ablation:}
Table~\ref{tab:ablation} shows that each component contributes to
NormLift. Removing refinement causes the largest drop, reducing mIoU by
$2.4$--$4.4$ points across label granularities. Within refinement,
removing reliability weights reduces 19-class mIoU by $1.3$ points, and
replacing mode selection with distance-weighted linear averaging reduces
it by $0.6$ points. These results support reliability-aware mode voting,
which selects a coherent neighbor feature instead of averaging nearby
CLIP features. Using only one factor of $R(j)$ also hurts performance:
the norm-only variant is unstable under single-view dominance, while the
view-diversity-only variant remains $1.7$ mIoU points below the full
score at 19 classes.

\paragraph{Runtime:}
Table~\ref{tab:timing} compares NormLift with SFS on LERF-OVS. Since both
methods share the same lifting stage, the difference comes from
post-lifting. NormLift takes $53.5$\,s on average for reliability
computation and mode voting, compared with $359.0$\,s for SFS, yielding
a $\mathbf{6.7\times}$ post-lifting speedup and a $2.0\times$ end-to-end
speedup ($295.9$\,s vs.\ $601.5$\,s). Peak GPU memory is essentially
identical between the two methods, since it is dominated by the shared
back-projection stage and the post-lifting operations of both methods
operate on aggregated per-Gaussian features. Appendix~\ref{app:runtime-details} shows more details.

\section{Conclusion}
\label{sec:conclusion}

We present \textbf{NormLift}, a training-free framework for feature
lifting in 3D Gaussian Splatting. By recasting per-Gaussian feature
lifting as a cosine alignment problem on the CLIP unit sphere, we show
that the standard back-projection rule together with $\ell_2$
normalization is exactly the closed-form optimum of a 3D-side
per-Gaussian objective, complementing existing rendering-side
analyses. From the same formulation, the back-projected norm
algebraically factors into intra- and inter-view consistency, and
calibrated by effective multi-view support yields a per-Gaussian
reliability score. This score guides a mode-based refinement that
respects the non-linearity of the CLIP feature space.

Experiments show that NormLift improves over prior
training-free and training-based baselines on ScanNet open-vocabulary
3D semantic segmentation, runs $\mathbf{6.7\times}$ faster than SFS in
the post-lifting stage at matched peak memory, and transfers
competitively to 2D rendering on LERF-OVS. Further analysis shows that
the reliability score is largely independent of opacity outside a
small zero-evidence regime, providing a per-Gaussian signal beyond
what visibility-based heuristics offer. We hope this 3D-side view of
feature lifting motivates future work that designs lifting objectives
directly in the geometry under which lifted features are used
downstream.

\paragraph{Limitations:}
(1) The framework takes the alpha-compositing weights $A_{ij}$ and the
underlying 3DGS geometry as fixed inputs, so geometric artifacts in the
reconstruction propagate into lifting.
(2) The lifted features inherit CLIP's own limitations on fine-grained
categories, compositional concepts, and objects outside its training
distribution.
(3) The mode-voting refinement requires at least some spatial neighbors
of an unreliable Gaussian to themselves be reliable; regions with
systematically unreliable neighborhoods cannot be recovered without
additional cues such as active view planning.

\begin{ack}
This work was supported by the King Abdullah University of
Science and Technology (KAUST) research fund under grant
BAS/1/1680-01-01.    
\end{ack}

\bibliographystyle{plainnat}
\bibliography{references}

\appendix

\section{Why a 3D-side formulation is needed}
\label{app:why-3d-side}

Sections~\ref{sec:back-projection} and the CLIP non-linear properties
highlight a mismatch between the standard back-projection rule and the
geometry of the feature space in which lifted semantics are ultimately
used. Back-projection produces an unconstrained Euclidean feature
$f_j$, while downstream CLIP-based queries compare unit-normalized
directions on the semantic sphere. This raises a basic question: should
feature lifting be understood as a rendering-side reconstruction problem,
or as a per-Gaussian semantic assignment problem on the CLIP sphere?

Most existing analyses take the rendering-side view. In this view,
Gaussian features are treated as variables that are linearly composed
through alpha blending to reconstruct 2D feature maps. This perspective
is useful because it mirrors the standard 3DGS rendering pipeline and
leads to tractable formulations. For example, existing methods justify
or implement feature lifting through weighted aggregation, inverse
rendering, or closed-form approximations to 2D reconstruction objectives.
However, this view has two limitations for open-vocabulary semantic
lifting.

The important distinction is not whether a linear combination appears
algebraically, but what is being modeled as linear. In rendering-side
feature reconstruction, the expression $\sum_j A_{ij}u_j$ is treated as
a rendered semantic feature, implicitly extending RGB alpha compositing
to CLIP-like embeddings. This assumes that semantic features can be
combined as Euclidean quantities and compared to a target 2D feature.

In our 3D-side formulation, the starting point is different. We do not
define a rendered semantic feature by linearly blending Gaussian
features. Instead, for each Gaussian, we aggregate scalar cosine
agreements $\langle u,B_i\rangle$ between a candidate unit direction
and its supporting 2D observations. The back-projected vector
$\sum_i A_{ij}B_i$ appears only after using the linearity of the inner
product:
\[
\sum_i A_{ij}\langle u,B_i\rangle
=
\left\langle u,\sum_i A_{ij}B_i\right\rangle .
\]
Thus, the linear combination is an algebraic sufficient statistic for
weighted semantic agreement, not a claim that interpolated CLIP
features are themselves valid semantic observations. This is why the
solution can still take the form of normalized back-projection while
the interpretation remains different from rendering-side linear
reconstruction.

\section{Per-Gaussian independence and the role of Property~3}
\label{app:per-gaussian-independence}

A key structural feature of Eq.~\eqref{eq:cosine-alignment} is that the
feature estimation problem decomposes across Gaussians. For a fixed
Gaussian $j$, solving for $u_j$ only involves its own rendering weights
$\{A_{ij}\}_{i=1}^{R}$ and the corresponding 2D semantic observations
$\{B_i\}_{i=1}^{R}$. It does not depend on the feature estimates
$u_k$ of other Gaussians $k\neq j$. In other words, Eq.~\eqref{eq:cosine-alignment}
defines a per-Gaussian semantic consensus problem rather than a joint
2D feature reconstruction problem.

Importantly, this per-Gaussian formulation does not assume that each
pixel is explained by a single Gaussian. A training observation $B_i$
may still contribute to multiple Gaussians whenever $A_{ij}>0$ for
several $j$. This faithfully reflects the standard alpha-compositing
process in 3DGS, where a rendered pixel may receive contributions from
many Gaussians along the same camera ray. The decoupling occurs only at
the feature estimation stage: once the geometry and alpha-compositing
weights are fixed, each Gaussian independently aggregates the semantic
observations assigned to it by its own rendering weights.

This is different from optimization-based feature rendering, where one
typically minimizes a global reconstruction objective of the form
\begin{equation}
  \min_{\{u_j\}}
  \sum_{i=1}^{R}
  \left\|
    B_i - \sum_{j=1}^{P} A_{ij} u_j
  \right\|_2^2 .
  \label{eq:global-feature-reconstruction}
\end{equation}
In such a formulation, the rendered feature
$\sum_j A_{ij}u_j$ couples all Gaussians that contribute to the same
pixel observation $i$. Consequently, the estimate of one Gaussian feature
depends on the estimates of other Gaussians sharing the same pixel/ray.
This coupling is natural for RGB reconstruction, where colors are
Euclidean quantities and alpha blending is physically meaningful. However,
it is less appropriate for CLIP-like semantic features, whose meaningful
representations lie on a non-linear semantic manifold.

Property~3 explains the main issue. CLIP-like features are compared by
their directions on the unit sphere, and semantically meaningful features
are not generally closed under arbitrary linear combinations. Therefore,
a linear mixture of features from distinct objects may move away from
either original semantic concept and drift off the semantic manifold.
The global reconstruction objective in Eq.~\eqref{eq:global-feature-reconstruction}
implicitly treats the rendered semantic feature
$\sum_j A_{ij}u_j$ as a valid Euclidean superposition of Gaussian
semantics. This assumption can become problematic when a pixel receives
non-negligible contributions from multiple semantically different
Gaussians.

Our formulation avoids this solver-side linear-superposition assumption.
Each Gaussian is assigned an independent unit direction on the CLIP
sphere:
\begin{equation}
  u_j^\star
  =
  \arg\max_{u\in S^{F-1}}
  \sum_{i=1}^{R} A_{ij}\langle u,B_i\rangle .
\end{equation}
Thus, the semantic identity of Gaussian $j$ is determined by the
contribution-weighted observations that support it, rather than by forcing
all Gaussians sharing a pixel to jointly reconstruct a single 2D feature
through a Euclidean sum. Other Gaussians still influence $u_j^\star$
indirectly through occlusion, depth ordering, and alpha compositing,
which are already encoded in the fixed weights $A_{ij}$. However, their
semantic feature estimates do not enter the optimization for $u_j^\star$.

We therefore distinguish two notions of coupling. At the data level,
observations remain shared: the same $B_i$ may provide evidence for
multiple Gaussians with different weights $A_{ij}$. At the solver level,
the semantic feature estimates are decoupled: each Gaussian extracts its
own semantically meaningful direction from the shared observations
without being constrained to participate in a linear reconstruction of
them. This distinction is central to our 3D-side interpretation of
feature lifting.

\section{Hyperparameter Sensitivity}
\label{app:refinement-hparams}

\begin{figure}[h]
\centering
\includegraphics[width=\linewidth]{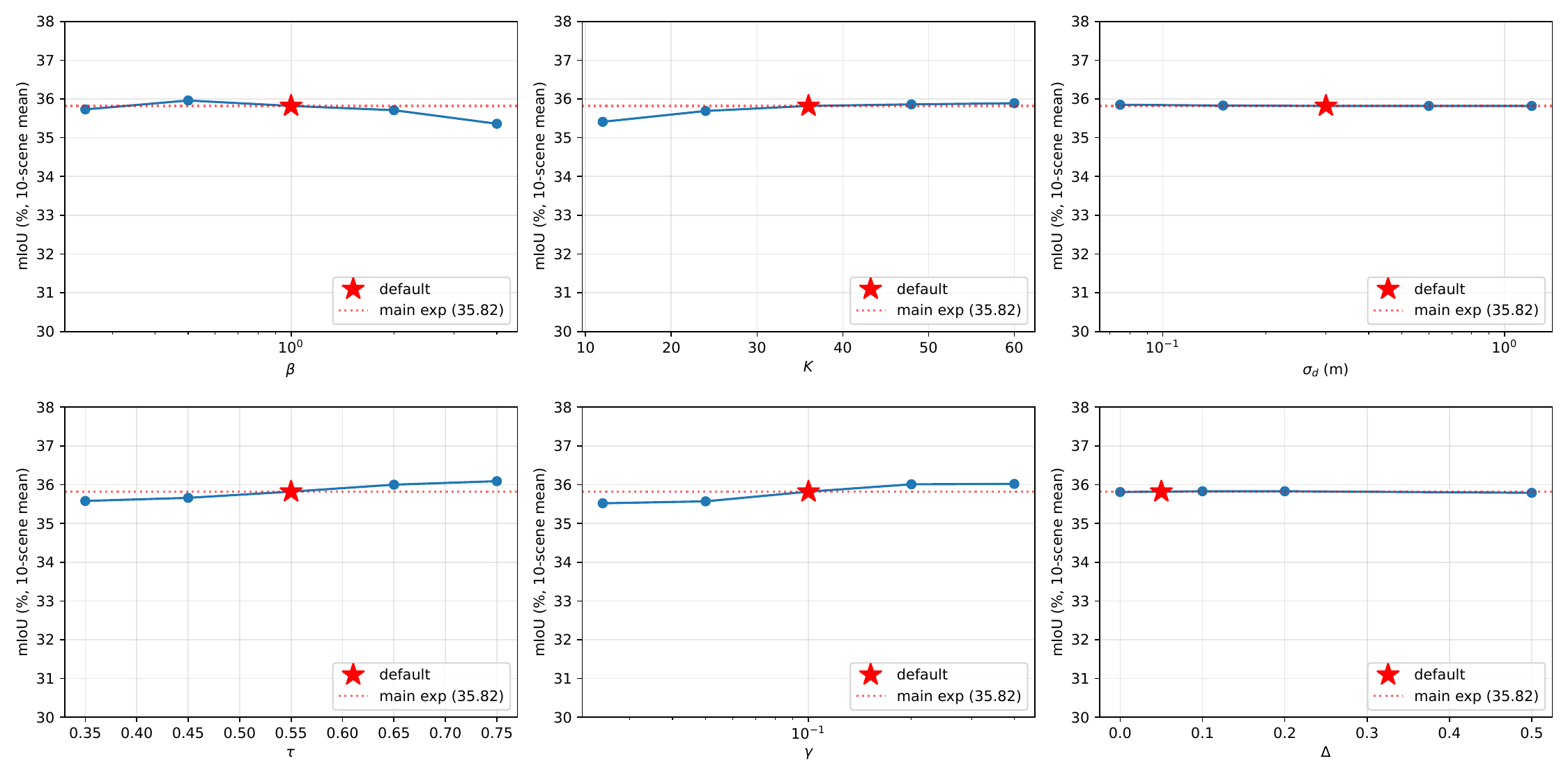}
\caption{Hyperparameter sensitivity on ScanNet (10-scene mean, 
19 classes). Each subplot shows mIoU as a function of one 
hyperparameter, with the others fixed at their defaults.}
\label{fig:hparam-sensitivity}
\end{figure}

We analyze the sensitivity of NormLift to its six hyperparameters:
the shrinkage strength $\beta$ in the reliability score
(Eq.~\ref{eq:reliability}), and the neighborhood size $K$, spatial
decay $\sigma_d$, semantic threshold $\tau$, gate sharpness $\gamma$,
and replacement margin $\Delta$ in the refinement procedure
(Eqs.~\ref{eq:support}--\ref{eq:margin}). For each hyperparameter we
vary its value across five points around the default ($\beta=1$,
$K=36$, $\sigma_d=0.3$\,m, $\tau=0.55$, $\gamma=0.1$, $\Delta=0.05$)
while keeping the others fixed. All sweeps use the 10 ScanNet scenes
of our main evaluation under the 19-class OpenGaussian protocol and
the released code. Results are shown in
Fig.~\ref{fig:hparam-sensitivity}.

\paragraph{Overall stability:}
Across all six hyperparameters and all 25 tested configurations,
NormLift's mIoU lies within $[35.36, 36.09]$, a range of $0.73$
points around the main-experiment result of $35.82$. No
hyperparameter exhibits a performance collapse within the tested
range, so NormLift does not require careful tuning. The defaults
were fixed before this sweep and are not the best values in it:
a larger $\tau$, $\gamma$ or $K$, or $\beta=0.5$, would each add
$0.1$--$0.3$ points. We keep the defaults unchanged.

\paragraph{Insensitive hyperparameters: $\sigma_d$ and $\Delta$:}
The spatial decay $\sigma_d$ changes mIoU by less than $0.05$ points
over $[0.075, 1.2]$\,m, and the replacement margin $\Delta$ by less
than $0.05$ points over $[0, 0.5]$. Both are expected. The candidate
set is already restricted to the $K$ nearest neighbours, so the
spatial kernel only re-weights points that are all close to the
target and the ranking of candidates is dominated by the reliability
and semantic-agreement terms; and the support score of the winning
candidate is rarely within $\Delta$ of the target's own score.

\paragraph{Saturating hyperparameters: $K$, $\tau$ and $\gamma$:}
The neighborhood size $K$ rises from $35.41$ at $K=12$ to $35.89$ at
$K=60$ and is flat within $0.1$ points above $K=36$. The semantic
threshold $\tau$ and the gate sharpness $\gamma$ increase
monotonically and saturate: $\tau$ moves from $35.58$ at $0.35$ to
$36.09$ at $0.75$, and $\gamma$ from $35.52$ at $0.025$ to $36.02$ at
$0.4$, with the defaults at the beginning of the plateau. Small
$\tau$ or $\gamma$ let semantically distant neighbours contribute to
the support score and cost up to $0.3$ points; beyond the default the
gate is already selective and further sharpening changes little.

\paragraph{Single-peaked hyperparameter: $\beta$:}
The shrinkage strength $\beta$ peaks at $\beta=0.5$ ($35.96$) and
falls off mildly on both sides, to $35.73$ at $0.25$ and $35.36$ at
$4$. An overly large $\beta$ over-shrinks the reliability of
Gaussians with few effective views and removes informative votes,
while a very small $\beta$ no longer discounts single-view dominance.

\section{Closed-form derivation of the per-Gaussian cosine alignment problem}
\label{app:closed-form-derivation}

We provide the full derivation of the closed-form solution stated in
Eq.~\eqref{eq:cosine-optimum}. The argument proceeds in three steps:
reduction by linearity, application of the Cauchy--Schwarz inequality,
and treatment of the boundary case.

\paragraph{Step 1: Reduction by linearity.}
By linearity of the inner product, the objective in
Eq.~\eqref{eq:cosine-alignment} can be rewritten as
\begin{equation}
\sum_{i=1}^{R} A_{ij}\langle u, B_i\rangle
=
\Bigl\langle u,\sum_{i=1}^{R} A_{ij}B_i\Bigr\rangle
=
\Bigl(\sum_{i=1}^{R} A_{ij}\Bigr)
\Bigl\langle u,
\frac{\sum_{i=1}^{R} A_{ij}B_i}{\sum_{i=1}^{R} A_{ij}}
\Bigr\rangle
=
W_j\langle u,f_j\rangle ,
\label{eq:app-linearity}
\end{equation}
where $W_j := \sum_{i=1}^{R} A_{ij}$ is the total rendering weight of
Gaussian $j$ and $f_j$ is the back-projected feature defined in
Eq.~\eqref{eq:backproj}. Since $W_j > 0$ is independent of $u$ and
does not affect the maximizer, the constrained problem reduces to
\begin{equation}
  \max_{u \in S^{F-1}} \langle u, f_j\rangle .
  \label{eq:app-reduced}
\end{equation}

\paragraph{Step 2: Cauchy--Schwarz.}
For any unit vector $u$, the Cauchy--Schwarz inequality yields
\begin{equation}
  \langle u,f_j\rangle \le \|u\|\,\|f_j\| = \|f_j\| ,
  \label{eq:app-cs}
\end{equation}
with equality if and only if $u$ is a non-negative scalar multiple of
$f_j$. Combined with the unit-norm constraint $\|u\|=1$, the unique
maximizer for $f_j \neq 0$ is $u = f_j / \|f_j\|$, and the optimal
value of the objective equals $\|f_j\|$.

\paragraph{Step 3: Boundary case.}
If $f_j = 0$, then $\langle u, f_j\rangle = 0$ for every unit vector
$u$, and the objective is identically zero on $S^{F-1}$. The
maximizer is therefore not unique in this degenerate case. This
``empty-evidence'' regime corresponds to Gaussians whose
contribution-weighted observations cancel out or whose total
rendering support is negligible. Such Gaussians are identified by the
reliability score $R(j) = 0$ in Sec.~\ref{sec:reliability}, and are
corrected by the mode-voting refinement procedure in
Sec.~\ref{sec:refinement}. We further characterize this regime
empirically in Appendix~\ref{app:reliability-opacity}, where we show
that it largely overlaps with very-low-opacity Gaussians.

\paragraph{Summary.}
Combining the three steps, the closed-form solution to
Eq.~\eqref{eq:cosine-alignment} is
\begin{equation}
  u_j^\star = \frac{f_j}{\|f_j\|}
  \quad (\text{for } f_j \neq 0),
  \qquad
  \text{with optimal value } \|f_j\| .
\end{equation}
Thus $u_j^\star$ coincides with the $\ell_2$-normalized back-projected
feature, and the optimal value of the cosine alignment objective is
exactly $\|f_j\|$. The latter identity is what makes $\|f_j\|$
meaningful beyond a normalization constant: it is itself the optimal
attained value of the per-Gaussian alignment objective, and we
revisit this fact when analyzing the structural meaning of the norm
in Sec.~\ref{sec:reliability}.

\section{Reliability score: basis, calibration and distribution}
\label{app:reliability-calibration}

\subsection{Basis of the reliability score}
\label{app:reliability-basis}
The reliability score in Eq.~\eqref{eq:reliability} is linear in
$\|f_j\|$. This appendix states the two facts that justify reading
$\|f_j\|$ as a measure of reliability. Note that $\|f_j\|$ is the norm
of the aggregated vector produced by back-projection, not of a CLIP
feature: each $B_i$ is a unit vector, so the magnitude of $f_j$ is
determined entirely by the aggregation.

\paragraph{Norm of a convex combination:}
By Eq.~\eqref{eq:backproj}, $f_j$ is a convex combination of unit
vectors, $f_j=\sum_i w_{ij}B_i$ with $w_{ij}\ge 0$ and
$\sum_i w_{ij}=1$. By the triangle inequality, $\|f_j\|\le 1$, with
equality if and only if all $B_i$ with $w_{ij}>0$ coincide, and
$\|f_j\|$ decreases as the observed directions spread apart. When the
rays observing Gaussian~$j$ fall on a single object, within and across
views, $\|f_j\|$ is close to one; when they straddle different semantic
regions, the directions partially cancel and $\|f_j\|$ shrinks. The
norm therefore measures how mixed the observations of a Gaussian are,
and Eq.~\eqref{eq:decomp} separates this mixing into its within-view and
across-view components. In directional statistics the same quantity is
the mean resultant length of the observed directions, the standard
estimator of the concentration parameter of a von Mises--Fisher
distribution~\citep{mardia2009directional}.

\paragraph{Attained value of the alignment objective:}
Appendix~\ref{app:closed-form-derivation} shows that the maximizer of
Eq.~\eqref{eq:cosine-alignment} is $u_j^\star=f_j/\|f_j\|$ and that the
maximum equals $\|f_j\|$. The direction and the norm are thus the two
outputs of the same problem: $u_j^\star$ is the single direction that
best explains the observations of Gaussian~$j$, and $\|f_j\|$ is how
well any single direction can explain them. Because $\|f_j\|$ is itself
the attained objective value, it enters Eq.~\eqref{eq:reliability}
linearly and requires no further transformation.

\paragraph{Calibration by effective views:}
$R(j)$ multiplies the norm by $N_{\mathrm{eff}}(j)/(N_{\mathrm{eff}}(j)+\beta)$.
The norm measures the agreement between the observations and their
consensus direction, and the second factor discounts norms that are
high only because a single view dominates the observations.
Appendix~\ref{app:trend} shows that this calibration is necessary: the
raw norm is non-monotonic in downstream accuracy, whereas $R(j)$ is
monotonic.

\subsection{The trend of accuracy and reliability}
\label{app:trend}
\begin{figure}
  \centering
  \includegraphics[width=\linewidth]{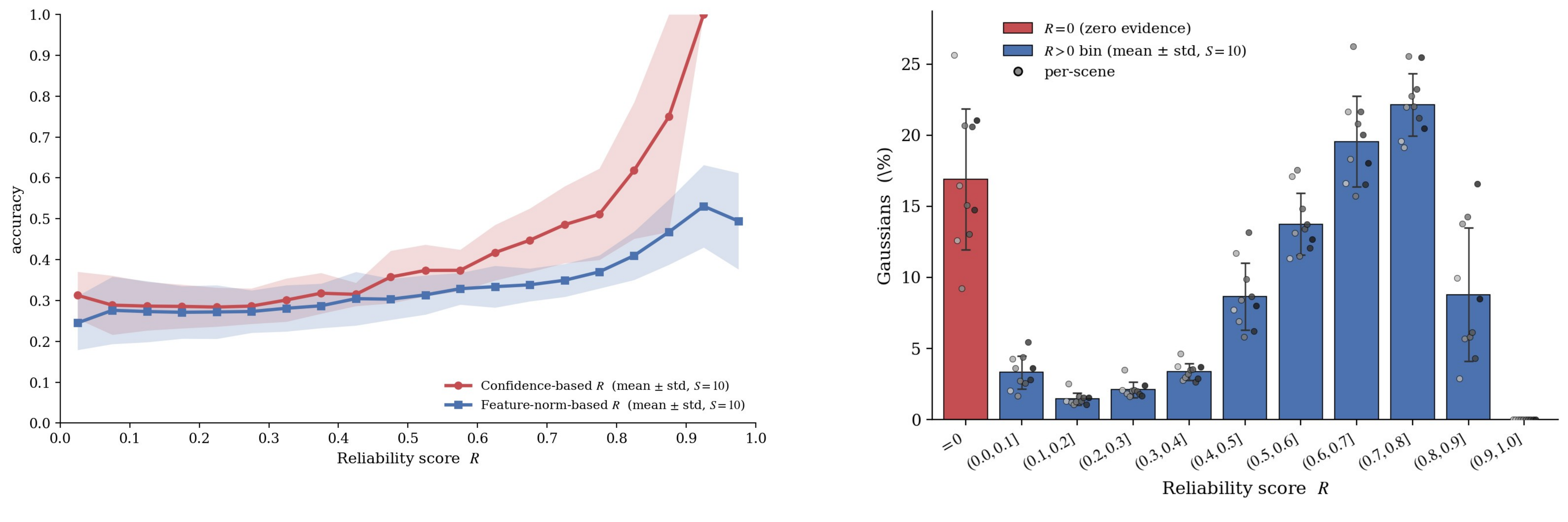}
\caption{
Reliability calibration and distribution on ScanNet.
Left: the proposed reliability score $R$ aligns more monotonically with downstream accuracy than the feature-norm-based score, showing the benefit of effective multi-view calibration.
Right: Gaussian distribution over reliability intervals, with zero-evidence Gaussians ($R=0$) shown in red and nonzero bins shown in blue.
The results indicate that $R$ provides an informative per-Gaussian signal for identifying unreliable primitives and guiding refinement.
}
  \label{fig:2forreliability}
\end{figure}

Figure~\ref{fig:2forreliability} left part illustrates why effective-view calibration is necessary. 
Using $\|f_j\|$ alone as a confidence signal produces a non-monotonic 
relationship with downstream accuracy: accuracy actually drops when 
$\|f_j\|$ approaches one. This counterintuitive behavior arises 
because a near-unit norm can be inflated by single-view dominance 
rather than genuine multi-view agreement, and Gaussians supported by 
only one dominant view are unstable in practice. Calibrating 
$\|f_j\|$ by $N_{\mathrm{eff}}$ removes this confound: the resulting 
reliability score $R(j)$ exhibits a monotonically increasing, 
approximately exponential relationship with accuracy, confirming 
that $R(j)$ is a far more faithful per-Gaussian confidence measure 
than $\|f_j\|$ alone.

\subsection{Reliability score versus opacity}
\label{app:reliability-opacity}

Figure~\ref{fig:2forreliability} right part shows there is an abnormal peak in the number of Gaussians at the point where the reliability is 0. A natural question for any per-Gaussian reliability measure is whether it
provides information beyond opacity-related signals, such as visibility
or rendering weight. This comparison is important because opacity-based
statistics have been used to remove noisy or weakly supported Gaussians
in prior work~\citep{occams}.

We analyze this relationship on the 10 ScanNet scenes used in our
experiments. We separate the Gaussians into two regimes. The first is the
zero-evidence subset
\begin{equation}
  \mathcal{G}_0 := \{j : R(j)=0\},
\end{equation}
which occurs when the back-projection accumulates no semantic evidence
for Gaussian $j$. This regime corresponds to a semantic observability
gap: the primitive exists in the 3D Gaussian representation, but receives
negligible semantic support from the training views. The second regime
contains the remaining Gaussians with $R(j)>0$.

\paragraph{Zero-evidence Gaussians align with opacity-based intuition:}
The zero-evidence subset $\mathcal{G}_0$ accounts for
$15.94\% \pm 4.14\%$ of all Gaussians and has sharply different physical
properties from the remaining Gaussians, as shown in
Table~\ref{tab:zero-evidence}. Its mean opacity is only $0.042$,
compared to $0.731$ for Gaussians with $R(j)>0$. Moreover, $86.32\%$ of
zero-evidence Gaussians have opacity below $0.1$, compared with only
$12.01\%$ in the nonzero-reliability regime. Their mean see-count is
also nearly zero ($0.112$ versus $11.86$). These statistics indicate
that $\mathcal{G}_0$ mostly consists of primitives that are barely
observed by the training views.

This physical distinction is also reflected semantically. The downstream
accuracy of $\mathcal{G}_0$ is substantially lower than that of the
remaining Gaussians ($35.61\%$ versus $46.21\%$). Thus, $R(j)=0$ is not
merely a numerical artifact: it identifies a class of primitives that
are both weakly observed and semantically unreliable. At this boundary,
the reliability score agrees with the intuition captured by opacity.

\begin{table}[t]
  \caption{Empirical characterization of the zero-evidence subset
  $\mathcal{G}_0$ versus the remaining Gaussians on ScanNet, averaged
  over 10 scenes. Zero-evidence Gaussians have much lower opacity,
  see-count, and downstream accuracy, indicating that $R(j)=0$ isolates
  a physically and semantically distinct regime.}
  \label{tab:zero-evidence}
  \centering
  \small
  \begin{tabular}{lccccc}
    \toprule
    Group & Ratio & Mean opacity & \% opacity $<0.1$ & Mean see-count & Acc \\
    \midrule
    All                       & $100\%$    & $0.622$ & $23.95\%$ & $9.85$  & $44.78\%$ \\
    $\mathcal{G}_0$           & $15.94\%$  & $0.042$ & $86.32\%$ & $0.112$ & $35.61\%$ \\
    $R(j)>0$                  & $84.06\%$  & $0.731$ & $12.01\%$ & $11.86$ & $46.21\%$ \\
    \bottomrule
  \end{tabular}
\end{table}

\paragraph{Beyond zero evidence, reliability is largely independent of opacity:}
The agreement between reliability and opacity does not extend to the
remaining $84.06\%$ of Gaussians with $R(j)>0$. Pooling
all Gaussians from this regime across all 10 scenes, the
rank-based correlations between $R(j)$ and opacity are essentially zero:
\[
  \rho_{\mathrm{Spearman}} = 0.002 \quad (p=0.31),
  \qquad
  \tau_{\mathrm{Kendall}} = 2\times 10^{-4} \quad (p=0.92).
\]
The overlap between low-reliability and low-opacity Gaussians is also
weak. The Jaccard overlap between the lowest-$20\%$ Gaussians ranked by
$R(j)$ and those ranked by opacity is $0.162$, which is below the
independence baseline of $0.20$. In addition, the mean opacity across
reliability quintiles is nearly flat:
\[
  0.56,\ 0.65,\ 0.63,\ 0.62,\ 0.60,
\]
from the lowest to highest reliability bins, showing no monotonic trend.

These results indicate that, once the zero-evidence boundary is excluded,
$R(j)$ and opacity capture different properties. Opacity reflects the
physical contribution or visibility of a Gaussian in rendering, whereas
$R(j)$ measures the semantic consistency and effective multi-view support
of the lifted feature.

\paragraph{Implication:}
The reliability score overlaps with opacity mainly at the boundary case
$\mathcal{G}_0$, where Gaussians receive almost no semantic evidence and
are also physically weakly observed. For the majority of Gaussians with
$R(j)>0$, opacity is nearly uninformative about semantic reliability. In
this regime, $R(j)$ provides an additional signal that is not captured by
opacity-based filtering. This distinction explains why our reliability
score can guide semantic refinement beyond simply removing low-opacity
or weakly visible Gaussians.

\section{ScanNet evaluation protocol and per-scene results}

\subsection{ScanNet evaluation protocol}
\label{app:scannet-protocol}
We follow the experimental protocol of OpenGaussian~\citep{opengaussian} 
for 3D semantic segmentation on ScanNet~\citep{dai2017scannet}. This 
appendix details the protocol for completeness.

For scene reconstruction, we initialize Gaussian Splatting from the raw 
scanned point clouds provided by ScanNet. Densification and position 
optimization are disabled, so all Gaussian primitives stay aligned with 
the input point cloud throughout reconstruction; only the remaining 
geometric and appearance attributes are optimized. This keeps the 
spatial layout of the Gaussians directly comparable to the ground-truth 
point cloud at evaluation time.

The features on 2D images follow the same process of LangSplat~\citep{qin2024langsplat} and we chose the L-level features.

Each method then constructs its semantic field on top of the 
reconstructed Gaussian scenes following its own design. At test time, 
every Gaussian primitive is classified by its open-vocabulary feature 
and compared against the ground-truth ScanNet point-cloud labels. We 
report mIoU and mAcc, providing a 3D point-level evaluation of semantic 
understanding.

We adopt the three label granularities defined by OpenGaussian. The 
19-class set covers the most common ScanNet object categories: 
\emph{wall, floor, cabinet, bed, chair, sofa, table, door, window, 
bookshelf, picture, counter, desk, curtain, refrigerator, shower 
curtain, toilet, sink, bathtub}. The 15-class subset removes 
\emph{picture}, \emph{refrigerator}, \emph{shower curtain}, and 
\emph{bathtub}; the 10-class subset further removes \emph{cabinet}, 
\emph{counter}, \emph{desk}, \emph{curtain}, and \emph{sink}.

OpenGaussian evaluates on 10 randomly selected ScanNet scenes: 
\texttt{scene0000\_00}, \texttt{scene0062\_00}, \texttt{scene0070\_00}, 
\texttt{scene0097\_00}, \texttt{scene0140\_00}, \texttt{scene0200\_00}, 
\texttt{scene0347\_00}, \texttt{scene0400\_00}, \texttt{scene0590\_00}, 
and \texttt{scene0645\_00}. We use the same set of scenes for all 
experiments.

\subsection{Per-scene results}
\label{app:per-scene}
NormLift is deterministic given the 3DGS reconstruction and the 2D
features: it involves no training, no random initialization and no
sampling, so repeated runs return identical numbers and run-to-run
error bars would be zero. The variation that does exist is across
scenes, and it reflects scene difficulty rather than the algorithm:
the same scenes are hard or easy for every method evaluated under
this protocol. We therefore report the results of Table~\ref{tab:scannet-3d}
for each of the ten ScanNet scenes (Table~\ref{tab:per-scene}) and
compare methods scene by scene. 
\begin{table}[h]
  \centering
  \caption{Per-scene results of NormLift on ScanNet (OpenGaussian
  protocol). The last row is the mean over the ten scenes and matches
  Table~\ref{tab:scannet-3d}.}
  \label{tab:per-scene}
  \small
  \begin{tabular}{lcccccc}
    \toprule
    & \multicolumn{2}{c}{19 classes} & \multicolumn{2}{c}{15 classes} & \multicolumn{2}{c}{10 classes} \\
    \cmidrule(lr){2-3}\cmidrule(lr){4-5}\cmidrule(lr){6-7}
    Scene & mIoU & mAcc & mIoU & mAcc & mIoU & mAcc \\
    \midrule
    0000 & 38.13 & 53.92 & 36.62 & 58.42 & 44.29 & 71.47 \\
    0062 & 34.81 & 68.57 & 34.63 & 68.63 & 53.33 & 82.53 \\
    0070 & 32.23 & 44.29 & 41.86 & 46.46 & 54.60 & 71.65 \\
    0097 & 39.63 & 59.29 & 37.44 & 64.74 & 53.29 & 72.70 \\
    0140 & 27.59 & 42.16 & 38.13 & 46.84 & 49.45 & 65.04 \\
    0200 & 38.89 & 53.30 & 41.37 & 58.07 & 39.06 & 55.11 \\
    0347 & 44.36 & 59.81 & 46.25 & 65.23 & 53.06 & 62.72 \\
    0400 & 32.97 & 58.25 & 33.75 & 62.63 & 40.52 & 67.16 \\
    0590 & 37.20 & 49.11 & 46.15 & 62.42 & 65.95 & 82.65 \\
    0645 & 32.42 & 52.75 & 37.59 & 59.27 & 42.53 & 63.04 \\
    \midrule
    Mean & 35.82 & 54.15 & 39.38 & 59.27 & 49.61 & 69.41 \\
    \bottomrule
  \end{tabular}
\end{table}

\section{Open-vocabulary 2D semantic segmentation}
\label{sec:exp2d}

We further evaluate NormLift on 2D open-vocabulary segmentation as a
complementary metric. Experiments are conducted on
LERF-OVS~\citep{kerr2023lerf,qin2024langsplat}, covering four scenes:
\textit{figurines}, \textit{ramen}, \textit{teatime}, and
\textit{waldo kitchen}. Following the evaluation protocol used in
SFS~\citep{sfs}, we render 2D feature maps from the lifted Gaussian
features and obtain object masks from text-query relevancy maps. All
methods use the same Gaussian geometry trained for 30k iterations, so
performance differences come only from the feature lifting strategy.

Table~\ref{tab:summary} compares NormLift with SFS~\citep{sfs} and
Occam's LGS~\citep{occams} using mIoU, mAcc, and locAcc. NormLift
achieves the best mean localization accuracy (0.807 versus 0.796 and
0.765) and is competitive in mIoU (0.511 versus 0.522 for Occam's LGS
and 0.502 for SFS). We discuss the mIoU gap explicitly, because it is
informative about what the method does and does not optimize.

\paragraph{Source of the mIoU gap:}
The mean gap to Occam's LGS is concentrated in a single
scene. On \textit{ramen} NormLift trails by 8.2 mIoU, whereas on
\textit{figurines}, \textit{teatime} and \textit{waldo kitchen} the
differences are $-0.5$, $+1.1$ and $+3.4$ points. This distribution is
consistent with Property~3. The drift described there is conditional:
it arises only when semantically distant features are blended with
comparable weights. Under the 2D protocol every pixel is an
alpha-composite of several Gaussians. On object interiors these
Gaussians belong to the same surface and the composite is a
small-angle mixture; at object boundaries the ray composites Gaussians
from different objects, which is the distant mixture that drifts.
\textit{Ramen} is the most cluttered of the four scenes, with many
small objects packed into a bowl, so a large fraction of its query
pixels lies near object boundaries, and a method that does not optimize
the composite is expected to lose most there.

\paragraph{Objectives of the two methods:}
Occam's LGS derives its estimator from the likelihood of reconstructing
the rendered 2D feature maps and adds a rendering-oriented filtering
step tuned to the composited output; its own ablation attributes a
sizable part of its accuracy to this filtering. NormLift optimizes the
coherence of each Gaussian's feature for independent 3D querying
(Section~\ref{sec:reliability}) and does not use the rendered
composite. The 2D protocol therefore evaluates NormLift outside the
setting it is designed for, and we report it as a transfer test in the
appendix rather than as a main result.

\begin{figure}
    \centering
    \includegraphics[width=\linewidth]{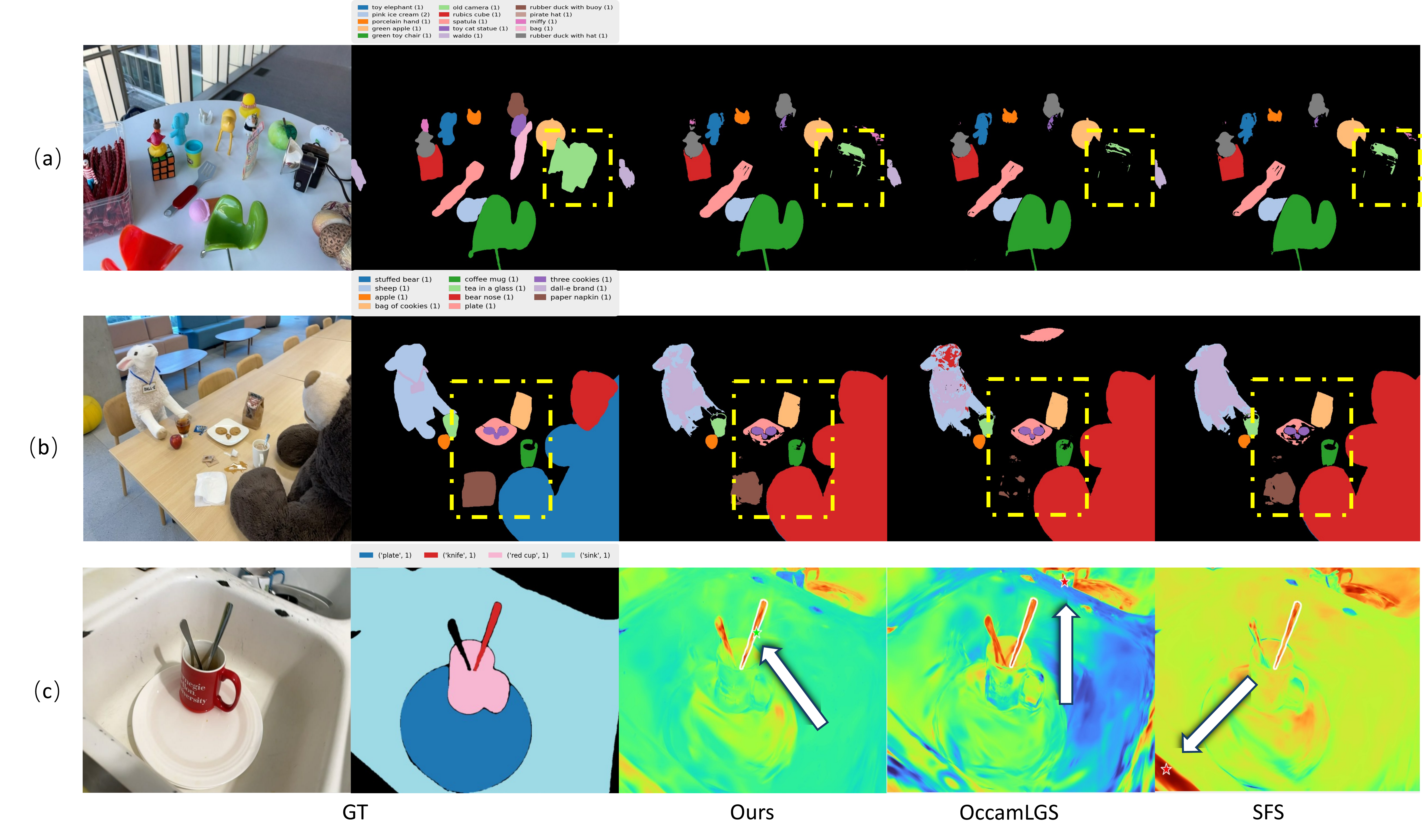}
    \caption{
    Qualitative comparison of semantic segmentation and localization across scenes. 
    (a,b) Segmentation results on \textit{figurines} and \textit{teatime}. 
    (c) Localization visualization on \textit{waldo kitchen}. 
    From left to right: input RGB image, ground-truth segmentation, SFS~\citep{sfs}, OccamLGS~\citep{occams}, and ours. 
    Our method produces comparable segmentation masks and yields more accurate and stable localization responses, as highlighted by the arrows.
    }
    \label{fig:2dseg}
\end{figure}

\paragraph{Localization accuracy:}
Localization accuracy measures whether the peak of the relevancy map
falls inside the queried object, and is the 2D metric most directly
tied to per-Gaussian semantic correctness. NormLift obtains the best
mean locAcc, in line with the quantity its objective targets.
Qualitative results in Fig.~\ref{fig:2dseg} show comparable
segmentation masks and more stable localization responses.

\paragraph{Dataset and protocol:}
We evaluate 2D open-vocabulary segmentation on the LERF-OVS
benchmark~\citep{kerr2023lerf,qin2024langsplat}, which contains four
real-world tabletop scenes: \textit{figurines}, \textit{ramen},
\textit{teatime}, and \textit{waldo kitchen}. Each scene provides
ground-truth polygon annotations for 8--17 object categories per
validation frame, with 22 validation frames in total. The scenes contain
objects with varying scale, density, and semantic ambiguity, making them
a challenging testbed for open-vocabulary feature lifting.

We follow the evaluation protocol used in the SFS~\citep{sfs} codebase.
For each method, we first render 2D feature maps from the lifted Gaussian
features and then obtain query-specific relevancy maps using CLIP text
queries. All methods use the same Gaussian geometry trained for 30k
iterations on each scene, ensuring that the comparison reflects the
feature lifting strategy rather than differences in reconstruction
quality.

\paragraph{Metrics:}
We report three metrics: mean Intersection-over-Union (mIoU), mean class
accuracy (mAcc), and localization accuracy (locAcc). For each category
$c$, let $\hat{M}_c$ denote the predicted mask obtained from the
relevancy map $a_c$, and let $M_c$ denote the ground-truth mask. We
compute
\begin{equation}
  \mathrm{mIoU}
  =
  \frac{1}{C}\sum_{c=1}^{C}
  \frac{|\hat{M}_c \cap M_c|}{|\hat{M}_c \cup M_c|},
  \qquad
  \mathrm{mAcc}
  =
  \frac{1}{C}\sum_{c=1}^{C}
  \frac{|\hat{M}_c \cap M_c|}{|M_c|}.
\end{equation}
Localization accuracy measures whether the peak response of the
relevancy map falls inside the ground-truth mask:
\begin{equation}
  \mathrm{locAcc}
  =
  \frac{1}{C}\sum_{c=1}^{C}
  \mathbf{1}\!\left[
    \arg\max_{(x,y)} a_c(x,y) \in M_c
  \right].
\end{equation}
Here, mIoU measures region overlap, mAcc captures per-class recall, and
locAcc evaluates whether the strongest semantic response is correctly
localized~\citep{kerr2023lerf}.

\paragraph{Mask thresholding:}
To convert a continuous relevancy map into a binary mask, we use the
same background-relative thresholding strategy for all methods. A pixel
is assigned to category $c$ if its relevancy exceeds the maximum
background relevancy by a fixed margin of $0.02$:
\begin{equation}
  \hat{M}_c(x,y)
  =
  \mathbf{1}\!\left[
    a_c(x,y)
    >
    \max_{b\in\mathcal{B}} a_b(x,y) + 0.02
  \right],
\end{equation}
where
\[
  \mathcal{B}
  =
  \{
  \text{``floor''}, \text{``wall''}, \text{``ceiling''},
  \text{``background''}, \text{``object''}, \text{``things''},
  \text{``stuff''}, \text{``texture''}
  \}.
\]
This thresholding rule is applied consistently to SFS, Occam's LGS, and
NormLift.

\begin{table}[t]
  \caption{Semantic segmentation and localization on LERF scenes.}
  \label{tab:summary}
  \centering
  \footnotesize
  \setlength{\tabcolsep}{3pt}
  \renewcommand{\arraystretch}{1.12}
  \begin{tabular}{lccc ccc ccc}
    \toprule
    & \multicolumn{3}{c}{mIoU $\uparrow$}
    & \multicolumn{3}{c}{mAcc $\uparrow$}
    & \multicolumn{3}{c}{locAcc $\uparrow$} \\
    \cmidrule(lr){2-4}\cmidrule(lr){5-7}\cmidrule(lr){8-10}
    Scene
    & SFS & Occam's & Ours
    & SFS & Occam's & Ours
    & SFS & Occam's & Ours \\
    \midrule
    Figurines
    & 0.569 & \textbf{0.593} & 0.588
    & 0.635 & 0.669 & \textbf{0.678}
    & 0.886 & 0.843 & \textbf{0.899} \\
    Ramen
    & 0.273 & \textbf{0.369} & 0.287
    & 0.458 & \textbf{0.632} & 0.500
    & 0.474 & \textbf{0.731} & 0.544 \\
    Teatime
    & 0.636 & 0.639 & \textbf{0.650}
    & 0.775 & 0.790 & \textbf{0.796}
    & \textbf{0.923} & 0.857 & 0.906 \\
    Waldo
    & \textbf{0.531} & 0.487 & 0.521
    & 0.695 & 0.703 & \textbf{0.720}
    & 0.777 & 0.753 & \textbf{0.877} \\
    \midrule
    Mean
    & 0.502 & \textbf{0.522} & 0.511
    & 0.641 & \textbf{0.699} & 0.674
    & 0.765 & 0.796 & \textbf{0.807} \\
    \bottomrule
  \end{tabular}
\end{table}

\begin{table}[h]
\centering
\caption{Wall-clock time and peak GPU memory on LERF-OVS, on a 
single NVIDIA RTX 4090. NormLift and SFS share the same lifting 
algorithm and thus the same lifting time per scene. The two 
methods differ in the post-lifting stages: SFS applies product 
quantization followed by refinement, whereas NormLift computes a 
reliability score and applies KNN mode-voting refinement.}
\label{tab:timing-detail}
\footnotesize
\setlength{\tabcolsep}{4pt}
\begin{tabular}{l l r r r r r r}
\toprule
Method & Scene & \#Gauss. & Lift (s) & \makecell{Rel. /\\Quant. (s)} & Refine (s) & Total (s) & Mem (GB) \\
\midrule
\multirow{5}{*}{SFS}
& figurines      &   951{,}644 & 449.24 &  99.53 & 280.00 & 828.77 &  7.48 \\
& ramen          &   479{,}180 & 131.02 &  25.78 &  92.17 & 248.97 &  4.71 \\
& teatime        & 1{,}929{,}418 & 189.45 & 343.38 & 188.95 & 721.78 & 15.16 \\
& waldo\_kitchen  & 1{,}537{,}735 & 200.25 & 225.47 & 180.62 & 606.34 & 12.09 \\
& \textbf{Mean}  & 1{,}224{,}494 & 242.49 & 173.54 & 185.44 & 601.46 &  9.86 \\
\midrule
\multirow{5}{*}{\textbf{NormLift}}
& figurines      &   951{,}644 & 449.24 &   3.21 &  31.59 & 484.04 &  7.48 \\
& ramen          &   479{,}180 & 131.02 &   0.54 &  10.29 & 141.85 &  4.71 \\
& teatime        & 1{,}929{,}418 & 189.45 &   0.80 & 100.40 & 290.65 & 15.17 \\
& waldo\_kitchen  & 1{,}537{,}735 & 200.25 &   0.90 &  66.10 & 267.25 & 12.09 \\
& \textbf{Mean}  & 1{,}224{,}494 & 242.49 &   \textbf{1.36} &  \textbf{52.10} & \textbf{295.95} &  9.86 \\
\bottomrule
\end{tabular}
\end{table}

\section{Details of run time comparison}
\label{app:runtime-details}

This appendix provides per-scene timing and memory details that 
supplement the runtime comparison reported in the main paper. All 
measurements are taken on a single NVIDIA RTX 4090 GPU, with one 
warm-up run followed by a timed run. CUDA synchronization is 
enforced before and after each measured stage.

\paragraph{Pipeline alignment.}
NormLift and SFS share the same lifting algorithm 
(rendering-weighted aggregation of CLIP features onto 3D 
Gaussians, Eq.~\ref{eq:backproj}) and therefore have identical 
lifting times per scene. The two methods diverge in the 
post-lifting stages. SFS performs product quantization to compress 
the lifted features into a codebook, followed by a refinement 
stage. NormLift instead computes a per-Gaussian reliability score 
$R(j)$ (Eq.~\ref{eq:reliability}) and applies KNN mode-voting 
refinement (Sec.~\ref{sec:refinement}). For a fair comparison, the 
reliability stage and the quantization stage are placed in the 
same column of Table~\ref{tab:timing} as the corresponding 
post-lifting bookkeeping operation in each method, even though 
they are not the same operation.

\paragraph{Per-stage timing.}
Table~\ref{tab:timing} reports the per-scene wall-clock time and 
peak GPU memory of NormLift and SFS on the four LERF-OVS scenes. 
The reliability score computation in NormLift takes only $1.36$\,s 
on average, two orders of magnitude faster than SFS's product 
quantization stage ($173.54$\,s). The KNN mode-voting refinement 
takes $52.10$\,s on average, $3.6\times$ faster than SFS's 
refinement ($185.44$\,s). Combining the two post-lifting stages, 
NormLift is $6.7\times$ faster than SFS in total post-lifting 
time, which translates to a $2.0\times$ end-to-end speedup 
($295.95$\,s vs.\ $601.46$\,s on average).

\paragraph{Memory.}
Peak GPU memory is essentially identical between NormLift and SFS 
across all four scenes (mean $9.86$\,GB, per-scene differences 
within $0.01$\,GB). This is expected because the dominant memory 
cost arises from loading the Gaussian parameters and the 
per-view CLIP feature maps in the lifting stage, which is 
shared by both methods. The post-lifting stages of both methods 
operate on aggregated per-Gaussian features and do not raise the 
peak. NormLift's speedup is therefore not obtained at the cost of 
additional memory.

\paragraph{Scene-level observations.}
Total runtime scales roughly with the number of Gaussians, but 
the post-lifting stage of SFS is more strongly affected by scene 
size than NormLift's. For example, on \textit{teatime} (1.9M 
Gaussians), SFS spends $532.33$\,s in post-lifting (quantization 
plus refinement), while NormLift spends only $101.20$\,s, a 
$5.3\times$ difference. On the smallest scene \textit{ramen} 
(0.5M Gaussians), the gap widens to $10.9\times$. This indicates 
that NormLift's per-Gaussian operations scale more favorably with 
scene complexity than SFS's quantization-based pipeline.

\section{Additional analyses}
\label{app:additional}

\subsection{Transfer of the reliability signal to another aggregator}
\label{app:geomedian}
To test whether $R(j)$ is tied to our voting operator, we replace
back-projection with the cosine-space geometric median of
VALA~\citep{wang2026visibility} under our frozen-Gaussian ScanNet protocol
(its in-rasterizer visibility gate does not transfer and is omitted),
and then apply our $R$-guided refinement with $R(j)$ computed as in
Section~\ref{sec:reliability}. Table~\ref{tab:geomedian} shows that
$R(j)$ improves the geometric-median features by $+2.38$ mIoU and
$+2.65$ mAcc, with all ten scenes improving. The signal therefore
transfers to a different robust scheme; the combination still trails
NormLift, whose aggregator is the one that attains the objective
$R(j)$ is derived from.

\begin{table}[h]
  \centering
  \caption{$R$-guided refinement on top of another aggregator
  (ScanNet, 19 classes, 10 scenes).}
  \label{tab:geomedian}
  \begin{tabular}{lcc}
    \toprule
    Scheme & mIoU & mAcc \\
    \midrule
    Spherical geometric median (adapted from VALA) & 32.96 & 50.53 \\
    \quad + $R$-guided refinement & 35.34 & 53.18 \\
    NormLift & 35.82 & 54.15 \\
    \bottomrule
  \end{tabular}
\end{table}

\subsection{Sensitivity to 3DGS geometry quality}
\label{app:geometry-quality}
We train 3DGS with densification on two ScanNet++ scenes
(\texttt{0d2ee665be}, small; \texttt{38d58a7a31}, large) and run
NormLift on checkpoints at 7k, 15k and 30k steps with unchanged
hyperparameters; since densified Gaussians are not aligned with the
ground-truth points, each point takes the features of its nearest
Gaussians. Results vary by less than one mIoU point across tiers
(Table~\ref{tab:geometry-quality}). Semantics is coarser than
geometry, so moderately displaced Gaussians keep their labels; weight
noise averages out over hundreds of observations; and floaters
produced by poor geometry receive inconsistent observations, fall into
the low-$R$ regime, and are repaired by the refinement.

\begin{table}[h]
  \centering
  \caption{NormLift under three quality tiers of the 3DGS
  reconstruction (ScanNet++, two scenes).}
  \label{tab:geometry-quality}
  \begin{tabular}{lcc}
    \toprule
    Training steps & mIoU & mAcc \\
    \midrule
    7k  & 19.25 & 34.29 \\
    15k & 19.86 & 34.80 \\
    30k & 19.26 & 34.59 \\
    \bottomrule
  \end{tabular}
\end{table}

\subsection{Generalization beyond ScanNet}
\label{app:generalization}
We ran
the same pipeline with all hyperparameters fixed to the ScanNet defaults
on ScanNet++ (large-scale indoor scenes, 100 classes, 15--34$\times$ more
Gaussians per scene) and HoliCity (outdoor street scenes), reproducing
SFS under the same protocol for comparison (Table~\ref{tab:generalization}).
NormLift improves over SFS on both datasets without per-dataset tuning.

\begin{table}[h]
  \centering
  \caption{Generalization with unchanged hyperparameters (mIoU / mAcc).}
  \label{tab:generalization}
  \begin{tabular}{lcc}
    \toprule
    Method & ScanNet++ & HoliCity \\
    \midrule
    SFS      & 19.14 / 25.53 & 37.92 / 43.92 \\
    NormLift & \textbf{21.93} / \textbf{30.74} & \textbf{40.47} / \textbf{46.30} \\
    \bottomrule
  \end{tabular}
\end{table}

\subsection{What the refinement edits}
\label{app:edit-audit}
Mixing can assign a Gaussian a semantics that none of its observations
carries, and neighbor voting could in principle overwrite correct
Gaussians in thin structures or overlapping objects. We therefore audited
every prediction change made by the refinement on ScanNet (19 classes,
10 scenes). Voting changes the prediction of 16.9\% of the labeled
Gaussians; among the changes that affect correctness, corrections
outnumber breakages by 1.67:1 (50.9k versus 30.6k), a net gain of $+2.18$
points in per-Gaussian accuracy (Table~\ref{tab:transition}). Stratifying
the labeled Gaussians by the fraction of their 36 nearest neighbors that
share their ground-truth class (Table~\ref{tab:stratified}), the net
change is positive in every group, including the 1.5\% of Gaussians with
fewer than a quarter of same-class neighbors, which is where thin
structures and overlapping objects fall. A foreign candidate replaces a
Gaussian only if it outscores the Gaussian's own direction under the
semantic gate $g$ and the margin $\Delta$ (Eq.~\ref{eq:margin}), so a
numerical majority of foreign neighbors does not translate into
rewriting power.

\begin{table}[h]
  \centering
  \caption{Before/after transition of voting edits (ScanNet, 19 classes,
  10 scenes, labeled Gaussians).}
  \label{tab:transition}
  \begin{tabular}{lcc}
    \toprule
    & before: correct & before: wrong \\
    \midrule
    after: correct & 349.2k & 50.9k \\
    after: wrong   & 30.6k  & 501.9k \\
    \bottomrule
  \end{tabular}
\end{table}

\begin{table}[h]
  \centering
  \caption{Accuracy change of the refinement, stratified by the fraction
  of same-class neighbors (ScanNet, 19 classes).}
  \label{tab:stratified}
  \begin{tabular}{lcc}
    \toprule
    Same-class neighbors & Share of Gaussians & Acc change \\
    \midrule
    0--25\%     & 1.5\%  & $+1.31$ \\
    25--50\%    & 10.0\% & $+2.79$ \\
    50--75\%    & 23.5\% & $+3.00$ \\
    $\geq 75\%$ & 65.0\% & $+1.81$ \\
    \bottomrule
  \end{tabular}
\end{table}

%%%%%%%%%%%%%%%%%%%%%%%%%%%%%%%%%%%%%%%%%%%%%%%%%%%%%%%%%%%%
\newpage
\input{checklist.tex}

\end{document}

%% file: checklist.tex
\section*{NeurIPS Paper Checklist}

%%% END INSTRUCTIONS %%%

\begin{enumerate}

\item {\bf Claims}
    \item[] Question: Do the main claims made in the abstract and introduction accurately reflect the paper's contributions and scope?
    \item[] Answer: \answerYes{} % Replace by \answerYes{}, \answerNo{}, or \answerNA{}.
    \item[] Justification: The abstract and introduction make four specific claims: (1) per-Gaussian feature lifting can be formulated as a cosine alignment problem on the CLIP unit sphere with a closed-form $\ell_2$-normalized back-projected feature as its optimum; (2) the norm of the back-projected feature factors algebraically into intra-view and inter-view consistency; (3) calibrated by effective multi-view support, this norm yields a per-Gaussian reliability score that is largely independent of opacity; (4) NormLift is training-free and improves over prior training-free and
training-based baselines on open-vocabulary 3D semantic segmentation on
ScanNet, ranking first on five of the six reported metrics and second on
15-class mIoU, while remaining competitive on LERF-OVS. Each claim is supported in: (1) Section~\ref{sec:cosine}; (2) 
Section~\ref{sec:reliability}; (3) Section~\ref{sec:reliability} and 
Appendix~\ref{app:reliability-opacity}; (4) Section~\ref{sec:exp3d} 
and Appendix~\ref{sec:exp2d}.

\item {\bf Limitations}
    \item[] Question: Does the paper discuss the limitations of the work performed by the authors?
    \item[] Answer: \answerYes{} % Replace by \answerYes{}, \answerNo{}, or \answerNA{}.
    \item[] Justification: We discuss the limitations of NormLift at the 
end of Section~\ref{sec:conclusion}, covering dependence on the 
underlying 3DGS geometry, inherited limitations of CLIP, the 
indoor-scene scope of our 3D evaluation, and failure modes of the 
mode-voting refinement when all spatial neighbors are unreliable.

\item {\bf Theory assumptions and proofs}
    \item[] Question: For each theoretical result, does the paper provide the full set of assumptions and a complete (and correct) proof?
    \item[] Answer: \answerYes{} % Replace by \answerYes{}, \answerNo{}, or \answerNA{}.
    \item[] Justification: The paper contains two theoretical results. (1) The closed-form solution to the per-Gaussian cosine alignment problem (Eq.~\ref{eq:cosine-optimum}) is proven in Appendix~\ref{app:closed-form-derivation} via a three-step argument (reduction by linearity, Cauchy--Schwarz inequality, and treatment of the boundary case $f_j=0$). All assumptions, including $A_{ij}\geq 0$, $\sum_i A_{ij}>0$, and $B_i\in S^{F-1}$, are explicitly stated. (2) The norm decomposition (Eq.~\ref{eq:decomp}) is presented as an algebraic identity rather than a theorem and is derived in full in Section~\ref{sec:norm-decomp}. The interpretation of $C_{\mathrm{intra}}$ and $C_{\mathrm{inter}}$ uses standard results from directional statistics (mean resultant length, \citep{mardia2009directional}). All formulas are numbered and cross-referenced.

    \item {\bf Experimental result reproducibility}
    \item[] Question: Does the paper fully disclose all the information needed to reproduce the main experimental results of the paper to the extent that it affects the main claims and/or conclusions of the paper (regardless of whether the code and data are provided or not)?
    \item[] Answer: \answerYes{} % Replace by \answerYes{}, \answerNo{}, or \answerNA{}.
    \item[] Justification: The paper provides the following information for reproducibility: (1) the closed-form lifting rule (Eq.~\ref{eq:cosine-optimum}), reliability score (Eq.~\ref{eq:reliability}), and refinement procedure (Eqs.~\ref{eq:support}--\ref{eq:margin}) are fully specified; (2) the evaluation protocol on ScanNet follows OpenGaussian~\citep{opengaussian} with three label granularities (Section~\ref{sec:exp3d}); (3) the 2D LERF-OVS protocol follows SFS~\citep{sfs} and is detailed in Appendix~\ref{sec:exp2d}, including the exact background-relative thresholding rule; (4) all CLIP features are extracted following the LangSplat~\citep{qin2024langsplat} pipeline at the $l$-level; (5) refinement hyperparameters are fixed across all experiments and listed in Appendix~\ref{app:refinement-hparams}; (6) compute environment is reported (3D experiments are on one NVIDIA Tesla V100 and 2D experiments are on one RTX 4090 GPU).

\item {\bf Open access to data and code}
    \item[] Question: Does the paper provide open access to the data and code, with sufficient instructions to faithfully reproduce the main experimental results, as described in supplemental material?
    \item[] Answer: \answerYes{} % Replace by \answerYes{}, \answerNo{}, or \answerNA{}.
\item[] Justification: The full implementation is released at
\url{https://github.com/kaust-violab/NormLift_release} under the MIT
license.

\item {\bf Experimental setting/details}
    \item[] Question: Does the paper specify all the training and test details (e.g., data splits, hyperparameters, how they were chosen, type of optimizer) necessary to understand the results?
    \item[] Answer: \answerYes{} % Replace by \answerYes{}, \answerNo{}, or \answerNA{}.
    \item[] Justification: NormLift is training-free and does not involve gradient-based optimization for feature lifting. The paper specifies: (1) the 3DGS geometry training schedule (30k iterations, following standard 3DGS practice); (2) the CLIP feature extraction protocol (LangSplat $l$-level features); (3) the ScanNet evaluation protocol (OpenGaussian, three granularities of 19/15/10 classes); (4) the LERF-OVS evaluation protocol (Appendix~\ref{sec:exp2d}); (5) all hyperparameters $(K,\sigma_d,\tau,\gamma,\Delta,\beta)$ for the reliability score and refinement, fixed across all scenes (Appendix~\ref{app:refinement-hparams}); (6) the background term set $\mathcal{B}$ used for relevancy thresholding. Hyperparameter sensitivity is analyzed in the appendix.

\item {\bf Experiment statistical significance}
    \item[] Question: Does the paper report error bars suitably and correctly defined or other appropriate information about the statistical significance of the experiments?
    \item[] Answer: \answerYes{} % Replace by \answerYes{}, \answerNo{}, or \answerNA{}.
\item[] Justification: NormLift is deterministic given the 3DGS geometry
and the 2D features (no training, random initialization or sampling), so
run-to-run error bars would be zero. The relevant variation is across
scenes, and Appendix~\ref{app:per-scene} reports per-scene results for all
ten ScanNet scenes under all three label granularities so that the spread
and the consistency of the gains can be assessed directly.


\item {\bf Experiments compute resources}
    \item[] Question: For each experiment, does the paper provide sufficient information on the computer resources (type of compute workers, memory, time of execution) needed to reproduce the experiments?
    \item[] Answer: \answerYes{} % Replace by \answerYes{}, \answerNo{}, or \answerNA{}.
    \item[] Justification: Section~\ref{sec:ablation} states that all experiments were conducted on one NVIDIA Tesla V100 GPU and one NVIDIA RTX 4090 GPU. Because NormLift is training-free, the computational cost is dominated by (a) the standard 3DGS training (30k iterations, comparable to all baselines) and (b) the per-scene lifting and KNN refinement, which complete within minutes per ScanNet scene. 
    
\item {\bf Code of ethics}
    \item[] Question: Does the research conducted in the paper conform, in every respect, with the NeurIPS Code of Ethics \url{https://neurips.cc/public/EthicsGuidelines}?
    \item[] Answer: \answerYes{} % Replace by \answerYes{}, \answerNo{}, or \answerNA{}.
    \item[] Justification: The research conforms with the NeurIPS Code of Ethics. The work uses publicly available datasets (ScanNet and LERF-OVS) under their respective licenses and does not involve human subjects, crowdsourcing, sensitive personal data, or models with high misuse risk.

\item {\bf Broader impacts}
    \item[] Question: Does the paper discuss both potential positive societal impacts and negative societal impacts of the work performed?
\item[] Answer: \answerYes{}
\item[] Justification: NormLift is a training-free framework for 
open-vocabulary 3D semantic segmentation. Positive applications 
include AR/VR scene understanding, robotic perception, and 3D content 
editing. Potential indirect negative impacts are limited to general 
considerations applicable to all 3D scene understanding research 
(e.g., downstream surveillance applications), which are not specific 
to our methodological contribution.

    
\item {\bf Safeguards}
    \item[] Question: Does the paper describe safeguards that have been put in place for responsible release of data or models that have a high risk for misuse (e.g., pre-trained language models, image generators, or scraped datasets)?
    \item[] Answer: \answerNA{} % Replace by \answerYes{}, \answerNo{}, or \answerNA{}.
    \item[] Justification: The paper does not release any pre-trained models, image generators, or scraped datasets. NormLift is a training-free post-processing framework operating on top of existing 3DGS reconstructions and pre-trained CLIP features, and presents no high-risk release.


\item {\bf Licenses for existing assets}
    \item[] Question: Are the creators or original owners of assets (e.g., code, data, models), used in the paper, properly credited and are the license and terms of use explicitly mentioned and properly respected?
    \item[] Answer: \answerYes{} % Replace by \answerYes{}, \answerNo{}, or \answerNA{}.
    \item[] Justification: All existing assets are properly credited via citations: 3D Gaussian Splatting~\citep{kerbl20233dgs}, CLIP~\citep{radford2021clip}, ScanNet~\citep{dai2017scannet}, LERF and LERF-OVS~\citep{kerr2023lerf, qin2024langsplat}, LangSplat~\citep{qin2024langsplat} (used for CLIP feature extraction), and the comparison baselines (OpenGaussian, LAGA, THGS, VALA, Occam's LGS, SFS, LUDVIG). ScanNet and LERF-OVS are used under their original research licenses; CLIP and 3DGS code are used under their respective open-source licenses.

\item {\bf New assets}
    \item[] Question: Are new assets introduced in the paper well documented and is the documentation provided alongside the assets?
    \item[] Answer: \answerYes{} % Replace by \answerYes{}, \answerNo{}, or \answerNA{}.
\item[] Justification: The only new asset is the released code
(\url{https://github.com/kaust-violab/NormLift_release}, MIT license),
documented by a README that describes the pipeline stages, data layout,
hyperparameters and usage. No new datasets or pre-trained models are
released.

\item {\bf Crowdsourcing and research with human subjects}
    \item[] Question: For crowdsourcing experiments and research with human subjects, does the paper include the full text of instructions given to participants and screenshots, if applicable, as well as details about compensation (if any)? 
    \item[] Answer: \answerNA{} % Replace by \answerYes{}, \answerNo{}, or \answerNA{}.
    \item[] Justification: This work does not involve crowdsourcing or research with human subjects. All evaluations are conducted on existing public datasets with pre-existing annotations.

\item {\bf Institutional review board (IRB) approvals or equivalent for research with human subjects}
    \item[] Question: Does the paper describe potential risks incurred by study participants, whether such risks were disclosed to the subjects, and whether Institutional Review Board (IRB) approvals (or an equivalent approval/review based on the requirements of your country or institution) were obtained?
    \item[] Answer: \answerNA{} % Replace by \answerYes{}, \answerNo{}, or \answerNA{}.
    \item[] Justification: This work does not involve human subjects research and therefore does not require IRB approval or equivalent review.

\item {\bf Declaration of LLM usage}
    \item[] Question: Does the paper describe the usage of LLMs if it is an important, original, or non-standard component of the core methods in this research? Note that if the LLM is used only for writing, editing, or formatting purposes and does \emph{not} impact the core methodology, scientific rigor, or originality of the research, declaration is not required.
    %this research? 
    \item[] Answer: \answerNA{} % Replace by \answerYes{}, \answerNo{}, or \answerNA{}.
    \item[] Justification: LLMs are not used as a component of the core methodology. CLIP~\citep{radford2021clip}, which is used as a vision-language foundation model for feature extraction, is a standard, widely-used component in the open-vocabulary 3DGS literature and not a novel methodological contribution of this paper. LLMs were used only for editing and drafting parts of the text and did not affect the methodology, experiments or conclusions.

\end{enumerate}